\documentclass[Afour,sageh,times]{sagej}

\usepackage{moreverb,url}

\usepackage[colorlinks,bookmarksopen,bookmarksnumbered,citecolor=red,urlcolor=red]{hyperref}
\usepackage[mathscr]{euscript}
\usepackage{graphicx}
\usepackage{float}
\usepackage{subfig}
\usepackage{caption}
\usepackage{algorithm}
\usepackage{algpseudocode}
\usepackage{adjustbox}
\usepackage{tabularx}
\usepackage{caption}
\usepackage{booktabs}

\usepackage{soul}

\newcommand\BibTeX{{\rmfamily B\kern-.05em \textsc{i\kern-.025em b}\kern-.08em
T\kern-.1667em\lower.7ex\hbox{E}\kern-.125emX}}

\def\volumeyear{2016}

\begin{document}

\captionsetup[figure]{name={Fig.}}

\runninghead{Smith and Wittkopf}

\title{Compliance while resisting: a shear-thickening fluid controller for physical human-robot interaction}

\author{Lu Chen\affilnum{1}, Lipeng Chen\affilnum{2}, Xiangchi Chen\affilnum{2}, Haojian Lu\affilnum{1}, Yu Zheng\affilnum{2}, Jun Wu\affilnum{1}, Yue Wang\affilnum{1}, Zhengyou Zhang\affilnum{2}, and Rong Xiong\affilnum{1}}

\affiliation{\affilnum{1}State Key Laboratory of Industrial Control Technology, Zhejiang University, Hangzhou, Zhejiang, China.\\
\affilnum{2}Tencent Robotics X Lab, Shenzhen, China.}

\corrauth{Yue Wang, Rong Xiong, the State Key Laboratory of Industrial Control Technology, Zhejiang University, Hangzhou, Zhejiang, China.}

\email{wangyue@iipc.zju.edu.cn, rxiong@zju.edu.cn}

\begin{abstract}
Physical human-robot interaction (pHRI) is widely needed in many fields, such as industrial manipulation, home services, and medical rehabilitation, and puts higher demands on the safety of robots. Due to the uncertainty of the working environment, the pHRI may receive unexpected impact interference, which affects the safety and smoothness of the task execution. The commonly used linear admittance control (L-AC) can cope well with high-frequency small-amplitude noise, but for medium-frequency high-intensity impact, the effect is not as good. Inspired by the solid-liquid phase change nature of shear-thickening fluid, we propose a Shear-thickening Fluid Control (SFC) that can achieve both an easy human-robot collaboration and resistance to impact interference. The SFC's stability, passivity, and phase trajectory are analyzed in detail, the frequency and time domain properties are quantified, and parameter constraints in discrete control and coupled stability conditions are provided. We conducted simulations to compare the frequency and time domain characteristics of L-AC, nonlinear admittance controller (N-AC), and SFC, and validated their dynamic properties. In real-world experiments, we compared the performance of L-AC, N-AC, and SFC in both fixed and mobile manipulators. L-AC exhibits weak resistance to impact. N-AC can resist moderate impacts but not high-intensity ones, and may exhibit self-excited oscillations. In contrast, SFC demonstrated superior impact resistance and maintained stable collaboration, enhancing comfort in cooperative water delivery tasks. Additionally, a case study was conducted in a factory setting, further affirming the SFC's capability in facilitating human-robot collaborative manipulation and underscoring its potential in industrial applications.

\end{abstract}


\maketitle

\section{1. Introduction}
\begin{figure} 
    \centering
    \includegraphics[width=2.9in]{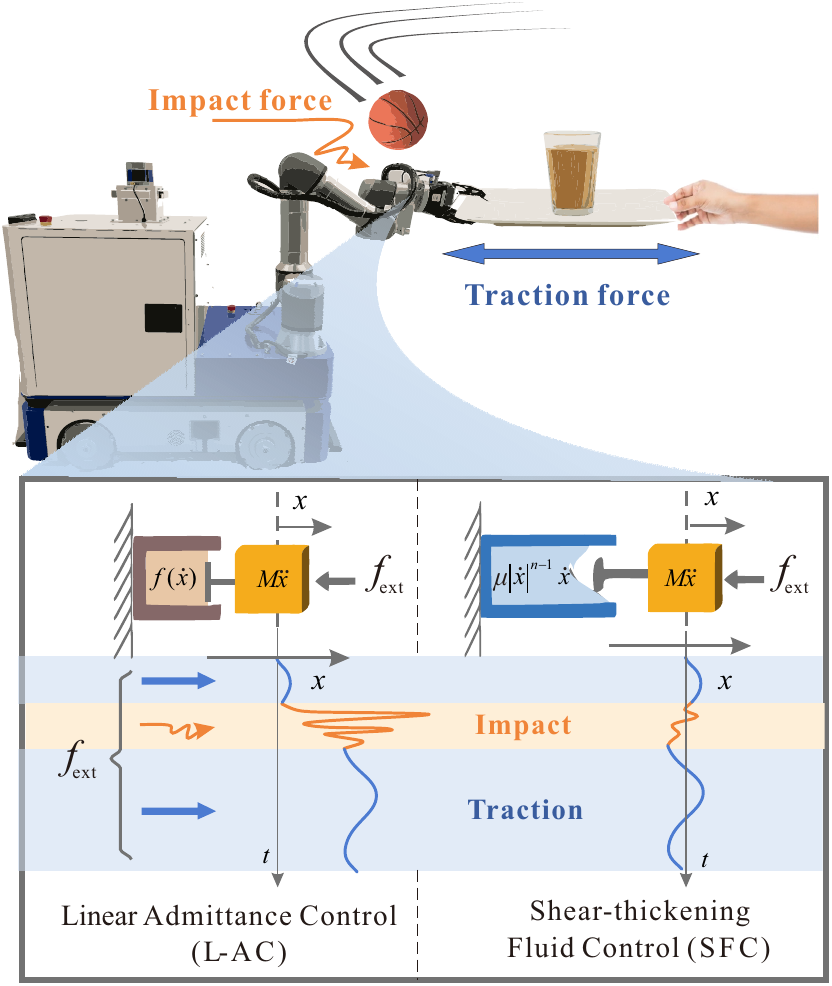}
    \caption{Interactive robot motion modeling and control based on shear-thickening fluid dynamics: the robot can comply with human traction and resist external impact}
    \label{fig:Sketch}
\end{figure}

Collaborative robots have increasingly become integrated into people's work and daily lives. Instead of confining robots to cages, there is now a growing demand for more direct physical interaction between humans and robots. Numerous collaborative robots have been developed using force and torque perception, employing impedance or admittance control for compliance. These robots are widely utilized in collaborative tasks, rehabilitation, and teleoperation scenarios~\cite{sun2019stability,spyrakos2019passivity,ferraguti2019variable}. However, in chaotic environments, the occurrence of external impacts is inevitable. As depicted in Figure~\ref{fig:Sketch}, if a ball suddenly strikes a robot while it is carrying an object alongside a person, the robot may struggle to differentiate between traction forces and impacts. This could lead to significant movement in the direction of the high-intensity impact, which may pose a safety risk to human-robot collaboration in close contact and can also affect task execution accuracy. Our goal is to enable the robot to demonstrate a new dynamic performance characterized by \textit{compliance with traction while resisting impacts}.

In the field of physical human-robot interaction (pHRI), impedance or admittance control is a standard method of achieving compliance interaction~\cite{hogan1984impedance}. 
Compared to impedance control, which heavily relies on accurate identification of robotic dynamic parameters~\cite{lammertse2004admittance}, admittance control only requires measuring contact forces using a force sensor and simulating virtual dynamics through kinematic closed-loop control. 
This makes admittance control more suitable for achieving compliant interactions in larger, non-backdrivable, or high-friction robotic systems~\cite{adams1999stable}. While admittance control ensures flexibility and stability in interactions (\cite{laghi2020unifying, aydin2021towards}), most admittance controllers are either linear or variable coefficient linear (\cite{ferraguti2019variable, aydin2018stable}). These linear admittance controllers have frequency responses that are independent of the input signal amplitude, which means they lack the ability to differentiate between traction and impact forces. There have been some studies on nonlinear admittance control, but they typically refer to cases where the controlled objects are nonlinear (\cite{morbi2013stability, bascetta2019ensuring, kim2019passivity}). It is uncommon for admittance controllers themselves to be \textit{nonlinear}. The existing research on nonlinear admittance control involve achieving convergence of assembly contact force (\cite{lai2014improving}) or predicting human intent (\cite{kang2019variable}), which do not align with the problem we are trying to address - specifically, complying with traction and resisting impact forces. Additionally, there is a lack of quantitative analysis for the dynamic response characteristics of nonlinear admittance controllers. Admittedly, there exist studies aimed at distinguishing between collision and intended contact forces to ensure collaborative safety (\cite{rodriguez2021human}). Nevertheless, the application of techniques such as mathematical model matching or signal threshold filtering (\cite{haddadin2017robot, lin2021adaptive}) can lead to time delays and recovery errors, thus impeding the real-time responsiveness and accuracy of human-robot interactions.

As mentioned by \cite{keemink2018admittance}, the essence of admittance control technique lies in the fact that ``by making the relation between the measured force and the reference velocity, \textit{the virtual model dynamics}, consistent with laws of mechanics, simulation of physical dynamical systems is possible." Inspired by this viewpoint, we realized that virtual dynamics can be extended to other physical processes, and admittance control can inherently incorporate nonlinear forms beyond the simulation of linear inertia-spring-damper systems. Shear-thickening fluids (STFs) exhibit a solid-liquid phase change characteristic: \textit{if you touch it gently, it will be as soft as water, but if you knock it hard, it will be as hard as a stone}. Leveraging the distinctive dynamic behavior of STFs, we propose a novel controller called Shear-Thickening Fluid Controller (SFC) within the realm of nonlinear admittance control. SFC allows robots to adapt to both traction and impact in pHRI, as shown in the Figure~\ref{fig:Sketch}. To ensure safety, we demonstrate the stability, passivity, and non-linear spatial velocity ratio of SFC. This ensures the robot's reliability when operating under various phases, including the traction phase, impact phase, and free motion phase. We also analyze the frequency domain response of SFC and derived its describing function, which provides insights into the dynamic characteristics of the robot under SFC. The discrete form of SFC for practical robot systems has also been presented. Guided by the theoretic analysis of the bandwidth, time constant, gain variation, and the discretization constraints, we further propose an automatic parameter tuning algorithm to determine appropriate controller parameter values based on dynamic response requirements in human-robot interaction scenarios, accounting for force requirements consistent with motion and impact limits as well as the system bandwidth of SFC. Finally, to evaluate the dynamic performance of SFC, we compare it with a linear admittance controller (L-AC) and a nonlinear admittance controller (N-AC) through simulation and real world experimentation. Our results show that, compared with L-AC, SFC exhibits superior impact resistance and performs effectively at low control frequencies. While both N-AC and SFC can suppress moderate strength impact disturbances, SFC demonstrates better impact resistance than N-AC under intense impact disturbances. Furthermore, in situations where traction and impact disturbances occur simultaneously, N-AC may suffer from self-excited oscillations, whereas SFC maintains stable collaboration while continuously inhibiting impact disturbances.
\begin{figure}[t] 
  \centering
  \includegraphics[width=2.8in]{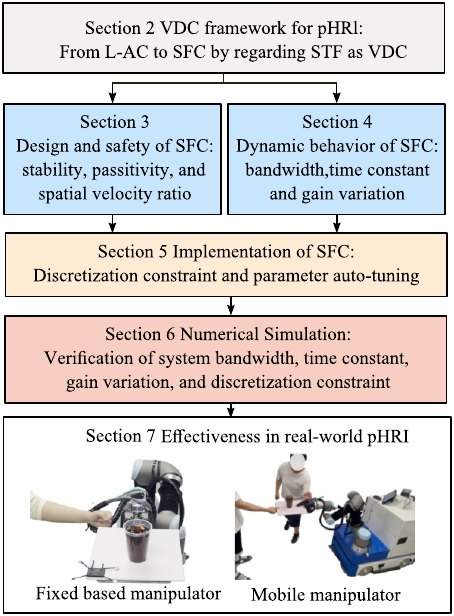}
  \caption{The structural relationship between the sections.}
  \label{fig:section_fra}
\end{figure}
This paper comprises several interconnected sections that provide a comprehensive overview of the proposed control framework for pHRI based on SFC, depicted in Figure \ref{fig:section_fra}. Section 2 introduces the virtual dynamics control based framework for pHRI and sets the foundation for the subsequent sections. Sections 3 and 4 focus on control design for SFC, as well as an analysis of its stability and dynamic characteristics. Section 5 provides an algorithm for auto-tuning the parameters of SFC for practical applications. Sections 6 and 7 validate SFC through simulations and physical experiments. Section 8 presents a case study on the application of the SFC in a factory setting. Section 9 provides a comprehensive overview and discussion of relevant studies in the field. 

\section{2. SFC: shear-thickening fluids as virtual dynamics in pHRI}
In this section, we present a pHRI framework based on virtual dynamics control and discuss the implementation of robot differential inverse kinematics. We clarify the structural differences between SFC and traditional linear admittance controllers. We propose to use SFC design to achieve different responses to traction and impact force. Additionally, we analyze the challenges of implementing SFC in robot virtual dynamics, which is the main focus of the subsequent chapters in this paper.

\subsection{2.1 The virtual dynamics control based framework for pHRI}

\begin{figure}[t]
    \centering \includegraphics[width=3 in]{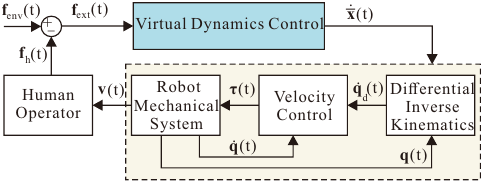}
    \caption{Control framework for the pHRI underlying virtual robot dynamics. The interaction model's solution with input $\mathbf{f}_\text{ext}$ yields the value of $\dot {\bar{\textbf{x}}}$, which the velocity-controlled robot needs to adhere to. This is achieved by computing the desired joint velocities $\dot{\textbf{q}}_\text{d}$ via differential inverse kinematics and regulating the joint torque $\boldsymbol{\tau}$ to enable the joint velocities $\dot{\textbf{q}}$ to track~$\dot{\textbf{q}}_\text{d}$.}
    \label{fig:framework}
\end{figure}

The pHRI control framework under virtual dynamics control is illustrated in Figure~\ref{fig:framework}.
Within this framework, the virtual dynamics control module is responsible for defining the desired interactive response characteristics of the robot. Using this virtual dynamics along with external forces from the environment, velocity commands for the robot can be generated, which serve as reference inputs for an inner-loop velocity controller, thereby driving the robot to achieve the desired interaction dynamic response.

\textbf{External force $\textbf{f}_\text{ext}$:} This combined force consists of the active force exerted by the environment or the operator on the robot, denoted as $\textbf{f}_\text{env}$, and the feedback interaction force, $\textbf{f}_\text{h}$, resulting from the interaction with human. Typically, this force is measured using sensors, such as six-axis force/torque sensors. As mentioned later, this paper abstracts the interaction between humans or the environment and the robot into force signals and differentiates between non-coupled, coupling transition, and coupled states of human-robot interaction through these force signals. External force furnishes direct information about the robot's interaction with the environment and accomplishes the outer-loop force control closed-loop.

\textbf{Virtual dynamics control:} This model describes the desired dynamic relationship between the robot and its environment. We believe that this relationship can be linear or nonlinear, typically based on physical principles. Depending on the specific application, different types of dynamic relationships can be chosen, such as damping, spring-damping, or mass-damping in the common admittance control framework. In this paper, we establish a virtual dynamics control based on shear thickening fluids, which will be described in detail later.

\textbf {Differential inverse kinematics:} After obtaining velocity commands in the Cartesian coordinate system, computing the inverse kinematics of the robot allows us to obtain the required joint space velocity commands. This step provides the joint velocity commands for the inner loop velocity controller while also dealing with issues such as singularities and redundancy regulation.

\textbf{Inner-loop velocity controller:} This is a controller based on velocity commands, responsible for generating joint torques to drive the robot. The velocity controller can be a traditional PID controller or other advanced control strategies. The task of the velocity controller is to make the actual velocity of the robot as close as possible to the velocity command generated by the virtual dynamic controller.

\textbf{Robot mechanical system:} Driven by the velocity controller, the robot attains the desired velocity and establishes contact with the human, eliciting a feedback force $\textbf{f}_\text{h}$ generated by the human operator.

In summary, the pHRI control framework based on robot virtual dynamics provides an effective tool for human-robot interaction. The design and execution of the Virtual Dynamics Control enable the robot to exhibit different dynamic characteristics during interaction. For instance, employing a linear inertia-damping model allows the robot to comply with human dragging. In this study, our goal is to achieve compliance with human traction and resistance to external impact through the utilization of a shear thickening fluids model. It is important to note that ensuring safety and flexibility during interaction requires specific design adjustments based on the requirements and interaction state.

\subsection{2.2 Differential inverse kinematics of robots}
 The direct kinematic mappings of interest for robots, can be written as

\begin{equation}
    \mathbf{x}_r(t) = \mathbf{f}\left( {\mathbf{q}\left( t \right)} \right)
\end{equation}
\begin{equation} \label{eq:kinematic}
    \dot{\mathbf{x}}_r\left( t \right) = \textbf{J}\left( {\mathbf{q}\left( t \right)} \right)\dot{\mathbf{q}}\left( t \right)
\end{equation}
where $\mathbf{q}(t)\in {\mathbb{R}^{{n_r}}}$ is the vector of joint variables, $\mathbf{x}_r\left( t \right) \in {\mathbb{R}^{{m_r}}}$ is a vector of task variables, and $\mathbf{f}(\mathbf{q}(t))$ is a differentiable nonlinear vector function with known structure and parameters for any robot, and $\textbf{J}(\mathbf{q}(t)) \in {\mathbb{R}^{{m_r} \times {n_r}}}$ is the configuration-dependent Jacobian matrix, formally defined as the partial derivatives of $\mathbf{f}(\mathbf{q}(t))$ with respect to $\mathbf{q}(t)$, denoted by ${{\partial f_i}/ {\partial q_j}},(i=1,...,m_r,j=1,...,n_r)$. The upper dot notation indicates a time derivative.

Jacobian matrix reflects the kinematic relationship of a robotic system, which is used to transform motion in Cartesian space to joint space. When the Jacobian matrix is not full rank, it means that the robot has singularities or redundancies at certain positions or configurations. Singularity occurrence leads to a sudden increase in joint velocities, even approaching infinity, resulting in unstable control. The presence of redundancy makes the inverse motion mapping non-unique. To address the issues of singularities and redundancies in inverse kinematics, we utilized a damped least squares method (\ref{eq:least_square}) to compute the pseudo-inverse of the Jacobian. When moving away from a singularity point, the application of the Moore-Penrose pseudoinverse is a method to solve for inverse kinematics while minimizing the 2-norm of joint velocities. The result is an optimal power configuration in a redundant configuration for control. When approaching singularities, we apply a small damping ($\lambda >0$) to reduce the impact of singularities. 
\begin{equation} \label{eq:least_square}
    \textbf{J}{\left( \textbf{q} \right)^\dag } = \left\{ \begin{array}{l}
\textbf{J}{\left( \textbf{q} \right)^\top}{\left( {\textbf{J}\left( \textbf{q} \right)\textbf{J}{{\left( \textbf{q} \right)}^\top}} \right)^{ - 1}},{\sigma _{\min }} > \varepsilon \\
\textbf{J}{\left( \textbf{q} \right)^\top}{\left( {\textbf{J}\left( \textbf{q} \right)\textbf{J}{{\left( \textbf{q} \right)}^\top} + \lambda \textbf{I}} \right)^{ - 1}},{\sigma _{\min }} \le \varepsilon 
\end{array} \right.
\end{equation}
where $\sigma_{\min}$ represents the minimum singular value of the Jacobian matrix, while $\varepsilon  > 0$ is a threshold value used to determine singularities.

Using the Jacobian pseudoinverse mapping (\ref{eq:least_square}), we can compute inverse kinematics to achieve a desired velocity in the task space.
\begin{equation}
    \dot{\mathbf{q}}_\text{d} = \textbf{J}{\left( \textbf{q} \right)^\dag }\dot {\bar {\mathbf{x}}}
\end{equation}
where $\dot {\bar {\mathbf{x}}}$ refers to the target velocity that corresponds to the virtual dynamics being considered, and $\dot{\mathbf{q}}_\text{d}$ is the configuration space velocity that achieves the $\dot {\bar{\mathbf{x}}}$.

The aforementioned kinematic mapping can also be employed to address redundant robot configurations. Both the fixed and mobile manipulators utilized in the subsequent experiments of this paper are controlled using this approach. It should be emphasized that this paper does not primarily focus on inverse kinematics, which is why there is only limited discussion on this topic. In the subsequent experiments, the focus is solely on the motion of robots in a fully-rank workspace, specifically when the rank of the Jacobian transpose is equal to the number of robot joints ($\text{rank}(\mathbf{J}\mathbf{J}^{\text{T}}) = n_r$).

Prior to the pHRI modeling, we must establish a foundational assumption regarding the robot system, in a manner similar to that done in \cite{ferraguti2019variable}. We anticipate that the low-level velocity controller within the robot system has undergone meticulous design and calibration to minimize tracking error and optimize dynamic response. This should enable the robot to accurately follow a feasible set-point. Notably, our primary emphasis is on the velocity state, with the set-point state referring to velocity.

\subsection{2.3 The pHRI control based on linear admittance control}
\begin{figure}[tbp]
	\centering
	\subfloat[]{\includegraphics[width=1.8in]{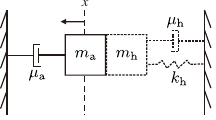}\label{fig.AC_pHRI}}\\
	\subfloat[]
 {\includegraphics[width=1.8in]{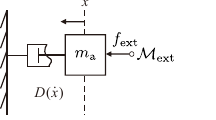}\label{fig.SFC_pHRI}}\\	
	\caption{Schematic view of physical human-robot interaction. (a) An admittance-controlled robot featuring virtual damping (${\mu _\mathrm{a}}$) and virtual mass (${m_\mathrm{a}}$), maintains contact with a human possessing inertia (${m_\mathrm{h}}$), stiffness (${k_\mathrm{h}}$), and damping (${\mu_\mathrm{h}}$).\\ (b) A robot that is controlled by SFC and equipped with STFs damping (${D(\dot x)}$) and virtual mass ($m$) is subjected to external environmental forces (${f_\mathrm{ext}}$).}
\end{figure}

When using admittance control to establish virtual dynamic control,  (as seen in ~\cite{keemink2018admittance}), a robot can be represented by its virtual dynamics with a rigid-body mass and some dissipation, as shown in Figure~\ref{fig.AC_pHRI}.
Without loss of generality, we will analyze one-dimensional motion in the following, as multi-dimensional motion is merely a simple superposition of one-dimensional motions.
The robot is then constrained to interact with its environment in accordance with a desired behavior, as specified below

\begin{equation} \label{eq:Admittance}
  {m_\mathrm{a}}\ddot x\left( t \right) + {\mu _\mathrm{a}}\dot x\left( t \right) = {f_\mathrm{ext}}\left( t \right)
\end{equation}
where ${m_\mathrm{a}}$ and ${\mu_\mathrm{a}}$ represent the desired virtual inertia and damping characteristics of the system, respectively.
We assumed ${m_\mathrm{a}}$ and ${\mu_\mathrm{a}}$ to be positive. ${f_\mathrm{ext}(t)}$ is an external force applied to the system, and is assumed to be measured by a force sensor that is attached to the wrist flange of the robot. The elastic portion of admittance control is utilized to attract the robot's end-effector to a desired position. However, since we aim to address the situation of a human operator manually controlling a robotic manipulator, this paper does not consider the elastic portion of the general admittance control model (\cite{siciliano2008springer}).

\subsection{2.4 The pHRI control based on SFC}
In light of the aforementioned reasons, this paper introduces a novel modeling approach as illustrated in Figure~\ref{fig.SFC_pHRI}. 
The robot is regarded as a separate controlled entity that exhibits desired dynamic characteristics based on virtual dynamics. Unlike linear admittance control, we replace the damping term with a nonlinear STFs form to establish the SFC.

\begin{equation} \label{eq:STFs}
    {m}\ddot x\left( t \right) + D\left( {\dot x}(t) \right) = {f_\text{ext}}\left( t \right)
\end{equation}
where ${\dot x}$ denotes the velocity of the controlled object, $m>0$ is a virtual inertial term, and $D\left( {\dot x} \right)$ is the nonlinear damping term, which is derived by analogy with the constitutive equation of STFs.For more detailed information regarding the specific form, please refer to Section 3. Equation (\ref{eq:STFs}) offers a non-linear damping term within an admittance control framework based on shear-thickening fluid dynamics. Meanwhile, equation (\ref{eq:Admittance}) represents the traditional linear admittance controller where the damping term is a linear function of velocity.  Admittance control can indeed include non-linear formulations (\cite{hogan1984impedance}). Therefore, we discuss our proposed SFC within the framework of admittance control.

\subsection{2.5 Problems statement}

When it comes to the environment or humans that interact with robots, the conventional approach often abstracts them as a second-order inertial-elastic-damping system (shown in dotted gray in Figure~\ref{fig.AC_pHRI}). During the analysis, it is commonly assumed that the interactive parties are fully coupled, meaning the robot and human or environment are in constant contact. In practical applications, the interaction between robots and the environment is not always in a tightly coupled state. For instance, when external objects impact with the robot or during the transition from non-coupled to coupled states (\cite{liberzon2003switching}). Therefore, instead of abstracting humans or the environment as a second-order dissipative system with specific constraints like EZ-width, we classify interaction forces based on contactless, traction, and impact forces, and then design controllers based on the magnitude and frequency characteristics of traction and impact forces. Please note that our study are not designing different controllers for different forces. Instead, we are designing one single SFC controller that can handle various interaction forces simultaneously, taking into account the characteristics of these forces. Specifically, we represent external forces ${f_{{\text{ext}}}}$ as a set ${\mathcal{M}_{{\text{env}}}}$, which have limited amplitude and frequency ranges, as explained below.
\begin{equation}
\begin{gathered}
  {\mathcal{M}_{{\text{env}}}} = \left\{ {\left. {{f_{{\text{ext}}}}\left( t \right)} \right|{\omega _{{\text{ext,min}}}} \leqslant \mathcal{W}\left( {{f_{{\text{ext}}}}\left( t \right)} \right) \leqslant {\omega _{{\text{ext,max}}}},} \right. \hfill \\
  \begin{array}{*{20}{c}}
  {}&{}&{} 
\end{array}\left. {\left| {{f_{{\text{ext}}}}\left( t \right)} \right| \leqslant {f_{{\text{ext,max}}}}} \right\} \hfill \\ 
\end{gathered}
\end{equation}
where $\mathcal{W}\left(  \cdot  \right)$ denotes the signal frequency computed using the fast-fourier transform, ${f_{{\text{ext,max}}}}$, ${\omega _{{\text{ext,min}}}}$, and ${\omega _{{\text{ext,max}}}}$ denote the upper limit of the amplitude and the lower and upper bounds of the frequency of the external force in the environment, with units of N and Hz, respectively. Then we can differentiate the interaction forces based on their magnitude and frequency ranges.  

\vspace{2mm} \noindent
\textbf{Definition 1: (Contactless)} 
A robot is said to be contactless when it experiences no external forces. Mathematically, this can be represented as
\begin{equation}
\begin{gathered}
  {{\mathcal{M}_{{\text{contactless}}}} = \left\{ {\left. {{f_{{\text{ext}}}}\left( t \right)} \right|\mathcal{W}\left( {{f_{{\text{ext}}}}\left( t \right)} \right) = 0,} \right.} 
 {\left| {{f_{{\text{ext}}}}\left( t \right)} \right| = 0} \} \hfill \\ 
\end{gathered}
\end{equation}
This also corresponds to a non-coupled state where there is no physical contact between the person and the robot.

\vspace{2mm} \noindent
\textbf{Definition 2: (Impact forces)}
If an external force acting on a robot exceeds a certain value $f_{\text{th}}$ or has a maximum frequency above a threshold value ($\varepsilon^+$), it is considered an impact force.

Mathematically, the condition for impact forces can be expressed as
\begin{equation} \label{eq:Force_Set}
\begin{gathered}
  {\mathcal{M}_{{\text{impact}}}} = \left\{ {\left. {{f_{{\text{ext}}}}\left( t \right)} \right|} \right. \hfill
{\omega _{{\text{ext,max}}}} \in \left( {{\varepsilon ^ + },\infty } \right) \cup \left. {f_\text{th} < {f_{{\rm{ext,max}}}}}  \right\}\hfill \\ 
\end{gathered}
\end{equation}
where ${{\varepsilon ^ + }}$ is the upper frequency limit of the external force ${f_\text{ext}}$, $f_{\text{th}}$ is the threshold magnitude for force

The actual impact force typically increases gradually to its maximum value over a short period of time and then decreases, thus, ${{\varepsilon ^ + }}$ may even approach infinity.

\vspace{2mm} \noindent
\textbf{Definition 3: (Traction forces)} If the external force acting on the robot is below a certain magnitude $f_\text{th}$ and has a maximum frequency below a threshold value ${{\varepsilon ^ - }}$ , it is considered as a traction force. Mathematically, this continuous force interaction can be expressed as:
\begin{equation} 
\begin{gathered}
  {\mathcal{M}_{{\text{traction}}}} = \left\{ {\left. {{f_{{\text{ext}}}}\left( t \right)} \right|} \right. \hfill
{\omega _{{\text{ext,max}}}} \in \left( {0,{\varepsilon ^ - }} \right) \cap \left. {{f_{{\rm{ext,max}}}} < f_\text{th}}  \right\}\hfill \\ 
\end{gathered}
\end{equation}
where ${{\varepsilon ^ - }}$ is the upper frequency limit of the external force ${f_\text{ext}}$. When ${{\varepsilon ^ - }}$ is sufficiently small, we can consider the interactive force between humans and robots to be continuous and smooth. During the traction force, the human and robot are typically in a coupled state.

\begin{figure}[t]
    \centering \includegraphics[width=2.5 in]{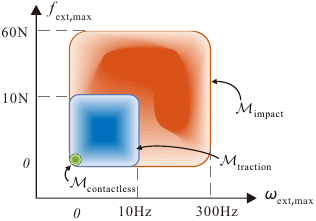}
    \caption{Sets of three kinds of interacting forces}
    \label{fig:differentForce}
\end{figure}

The amplitude and frequency range are calculated based on the actual conditions of the human or environment. For example, when human is pulling a robot, the exerted traction force is limited to $(f_{\text{ext,max}}, 0$, $\omega_{\text{ext,max}}) = (10, 0, 10)$. In instances when a robot is subjected to an impact force in its environment, the set of external forces is given by $(f_{\text{ext,max}}, 0, \omega_{\text{ext,max}}) = (60, 0, 300)$. These ranges align with real-world contact interactions. As shown in Figure \ref{fig:differentForce}, it depicts the relative relationship among three sets of forces. The impact force set includes components of traction forces, but its energy is predominantly concentrated in the high-frequency and high-amplitude range. On the other hand, the energy of the traction force set is concentrated within a smaller range of moderate amplitudes and frequencies. Our goal is to design an SFC controller that can suppress the high-frequency and high-amplitude portion of the impact force set while still complying with the traction forces. It is important to note that our proposed SFC does not employ adaptive or segmented control methods. Instead, we utilize the nonlinearity of the controller itself to achieve the desired effects using a fixed set of parameters.

\textbf{Problem:} Our objective was to design the form of SFC by incorporating the dynamic characteristics of shear-thickening fluids. This controller aims to effectively comply with traction forces ($\mathcal{M}_{\text{traction}}$) while withstanding impact forces ($\mathcal{M}_{\text{impact}}$) and ensuring stability during the interaction process. Furthermore, we need to propose a method for determining the parameters of the SFC based on the amplitude and frequency domain characteristics of the traction forces ($\mathcal{M}_{\text{traction}}$) and impact forces ($\mathcal{M}_{\text{impact}}$).

\subsection{2.6 Issues in designing virtual dynamics with~SFC}
The mapping between external force perception and velocity control commands generated by the virtual dynamics module, it is often referred to as an interactive controller, with admittance control being the most typical case~\cite{keemink2018admittance}. In this paper, we propose a new design for the robot virtual dynamics module in the human-robot interaction framework, aiming to endow the robot with dynamic performance in shear-thickening fluids. However, to make this idea feasible, several key technical challenges need to be addressed:

\begin{itemize}
\item{
\textbf{Controller design:} For an SFC controller of the type given in Equation (\ref{eq:STFs}), a non-linear damping term, $D\left( {\dot x} \right)$, needs to be designed such that the controller can abstractly express the physical characteristics of shear-thickening fluids and achieve a feasible robot virtual dynamic model (Section 4).
}

\item{
\textbf{Safety assurance:} It is necessary to prove that the controller composed of the virtual dynamics model is stable during free movement, human-robot interaction, interaction transitions, and impact disturbances to ensure the safety of both the robot and the human (Section 4).
}

\item{
\textbf{Dynamic behavior analysis:} Quantitative analysis of dynamic behavior helps understand SFC responses with the impact and traction forces. Frequency domain analysis reveals controller response characteristics under different frequencies and amplitudes. However, performing frequency domain analysis on nonlinear systems is difficult due to the limitations of Laplace Transform. Approximate methods like describing function are used instead (\cite{slotine1991applied}). Solving the nonlinear equation in the SFC system is challenging due to the presence of trigonometric functions and higher-order powers. Overcoming this challenge is essential for analyzing SFC dynamics and guiding controller design (Section 5).}

\item{
\textbf{Discretization constraint:} When the controller is used for a robot system, it is implemented on a discrete digital system. Therefore, it is necessary to analyze the constraints imposed by numerical integration on the system bandwidth and sampling time of the virtual dynamics module, in order to ensure that the system does not undergo vibrations (Section 6).
}

\item{
\textbf{Parameter design guidelines:} Based on the theoretical basis analyzed above, specific steps for controller's parameter design need to be given according to the actual application to meet the requirements of desired pHRI dynamic characteristics (Section 6).}
\end{itemize}

\section{3. The design and safety of the SFC}

Shear-thickening fluids (STFs) are a class of non-Newtonian fluids that have gained significant attention for their unique rheological properties and potential applications in various industrial fields~\cite{nrelated2}. In particular, STFs have been used in protective gear to enhance impact resistance and reduce the risk of injury to the wearer~\cite{nrelated5}. Unlike Newtonian fluids, STFs exhibit a stress-dependent viscosity, where the viscosity changes under external forces. This phenomenon is attributed to the transition between solid and liquid states under mechanical stress~\cite{nrelated1}. As such, the dynamic properties of STFs are highly attractive for developing controllers that are both impact-resistant and compliant.

In this section, we propose a novel controller inspired by the constitutive equation of STFs, which we refer to as the Shear-Thickening Fluid Controller (SFC). Our design leverages the unique properties of STFs to create controllers that can adapt to sudden changes in external forces, while remaining safe and compliant for use in human-robot interaction. 
We conduct a thorough analysis of the SFC's stability, passivity, and spatial velocity ratio, demonstrating its safety and reliability and highlighting its dynamic properties that distinguish it from linear controllers.

\subsection{3.1 Design of the SFC}

\begin{figure}[t]
\centering
\includegraphics{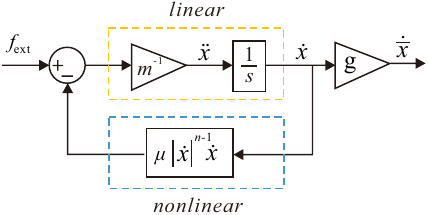}  
\caption{Block diagram of a SFC. A mapping relationship between force and velocity is described, corresponding to the virtual dynamics module shown in Figure~\ref{fig:framework}.}
\label{fig:Block_diagram}
\end{figure}

The proposed Shear-Thickening Fluid Controller (SFC) is defined by the following equation:

\begin{equation}\label{eq:with_mass}
\left\{ \begin{gathered}
  m\ddot x + \mu {\left| {\dot x} \right|^{n - 1}}\dot x = {f_\text{ext}} \hfill \\
  \dot{ \bar {x}} = g\dot x \hfill \\ 
\end{gathered}  \right.
\end{equation}
where, $f_\text{ext}$ represents the external force applied to the object, and ${\dot x}$ denotes the velocity of the controlled object. $m>0$ is a virtual inertial term, $g>0$ is a gain term used to regulate the system output, and $\dot{ \bar {x}}$ is the system output.
$\mu \ge 0$ is the apparent viscosity, a material characteristic describing the internal flow resistance. $n>1$ is a power-law term for shear thickening fluids, introducing the nonlinear characteristics. $D\left( {\dot x} \right) = \mu {\left| {\dot x} \right|^{n - 1}}\dot x$ is the nonlinear damping term, which is derived by analogy with the constitutive equation of STFs. Please refer to Appendix B for a detailed derivation.

The SFC block diagram is presented in Figure~\ref{fig:Block_diagram}, with the following design features.

\begin{figure}[t]
\centering
\includegraphics{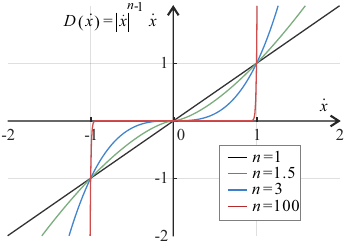}  
\caption{Non-linear damping force behavior in relation to velocity.}
\label{fig:nonlinearDamping}
\end{figure}

\textbf{Non-linear damping term:} $\mu {\left| {\dot x} \right|^{n - 1}}\dot x$, is derived from the constitutive equation of shear-thickening fluids (STFs), which allows the controller to emulate the dynamic properties of STFs. This feature provides the SFC with the ability to adjust its viscosity in response to external forces, leading to improved impact resistance and compliance. Specifically, the non-linear damping term increases with the object's velocity, thereby dissipating kinetic energy and enhancing impact resistance when the external force is strong. Conversely, when the external force is weak, the damping can be low, resulting in improved compliance and flexibility. Figure \ref{fig:nonlinearDamping} shows the relationship between non-linear damping force (y-axis) and velocity (x-axis).The non-linear damping equation used is $D\left( {\dot x} \right) = \mu {\left| {\dot x} \right|^{n - 1}}\dot x$, with $\mu$ set to 1, and n taking values of 1, 1.5, 3, and 100. The curve of $D(\dot x)$ is monotonically increasing and symmetric about the origin. For velocity $0 < \dot x < 1$, the non-linear damping curves with n = {1.5, 3, 100} are lower than the linear damping curve (n = 1), making it easier to traction the robot. At higher velocities $1< \dot x$, the non-linear damping curves with n = {1.5, 3, 100} are above the linear damping curve. This results in rapid dissipation of energy when moving away from the equilibrium point ($\dot x = 0$). Increasing n enhances the non-linear characteristics and improves system compliance and resistance. However, a high value of $n$ (e.g., $n=100$) can lead to an ill-conditioned system due to an extremely steep damping curve (See Appendix G for details). Therefore, it is sufficient for $n$ to be adequate; higher values are not necessarily better.

\textbf{Inertial term:} $m$, plays a crucial role in the SFC controller. This is because achieving zero inertia across the entire frequency range is impossible, as the force-velocity relationship leads to division by zero, as noted in Keemink's work~\cite{keemink2018admittance}. Additionally, by incorporating the inertial term, the velocity control command can be obtained through the integration of acceleration, which helps to alleviate control discontinuity and ensures smoother system motion. Furthermore, most mechanical systems in the real world have non-negligible mass, resulting in delayed control and actuation characteristics. Therefore, including the inertial term allows for a better representation of the physical phase shift observed in real systems.

\textbf{Gain term:} $g$, is added at the output to adjust the system's overall gain. The amplitude-frequency response curve can be translated longitudinally as a whole in the Bode diagram. The gain term $g$ does not affect the system bandwidth and can be ignored, so we set it to 1 in subsequent analyses.

 \subsection{3.2 Safety of the SFC}
By analyzing the non-linear spatial velocity ratio, stability and passivity properties of the SFC, safety of the robot under SFC in different interaction phases can be ensured, including but not limited to free motion, traction and impact phases. The main result on the safety is stated below.

\vspace{2mm} \noindent
{\bf{Theorem 1:}} Consider a controller be of the form (\ref{eq:with_mass}), with $m>0, \mu>0, n>1$. Then, the controller exhibits the following three properties: 

\begin{figure}[!t]
\centering
\subfloat[]{\includegraphics[width=2in]{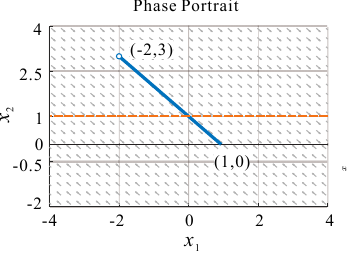}%
\label{fig:SNNFC_P_n1}}
\hfil
\subfloat[]{\includegraphics[width=2in]{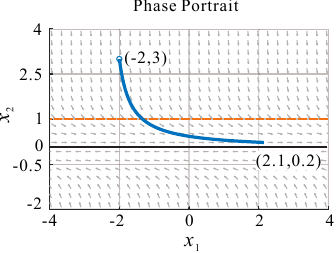}%
\label{fig:SNNFC_P_n3}}
\caption{The phase trajectory comparison of AC and SFC ($\mu =1, M=1$). (a) L-AC ($n = 1$). (b) SFC ($n=3> 1$).}
\label{fig:phase_trajectory}
\end{figure}

\vspace{2mm} \noindent \textbf{Property 1: (Non-linear spatial velocity ratio).} The SFC's phase trajectory is plotted on the phase plane with $x_1$ representing the position state on the horizontal axis and $x_2$ representing the velocity state on the vertical axis. The spatial velocity ratio ${R_\text{sv}}$ is defined as the ratio of the second state variable $x_2$ to the first state variable $x_1$ at a specific point on a phase trajectory, expressed as ${R_\text{sv}} \triangleq \frac{{{\text{d}}{x_2}}}{{{\text{d}}{x_1}}}$. Then, the SFC's spatial velocity ratio exhibits a nonlinear form, given by the following equation:
\begin{equation} \label{eq: dx2/dx1_1}
{R_\text{sv}} = - \frac{\mu }{m}{\left| {{x_2}} \right|^{n - 1}}
\end{equation}

\begin{proof}
For detailed analysis, see Appendix C
{\hfill $\blacksquare$\par}
\end{proof} 

The nonlinear spatial rate ratio primarily characterizes the convergence behavior of a robot's position and velocity in a \textbf{free motion phase}, i.e., in the absence of external forces, when the robot's velocity deviates from a fixed point and exhibits free motion. This ratio also reflects the robot's stability in the non-coupled state. The nonlinear spatial rate ratio property implies that the robot with SFC is capable of achieving rapid convergence under significant disturbances and maintaining compliance under small deviations, as it can adaptively adjust the virtual damping based on its state.

For instance, consider two SFC systems expressed by the equation (\ref{eq:with_mass}), wherein both systems exhibit equivalent positive values of $M$ and $\mu$. One of the systems has a degree of nonlinearity $n>1$, while the other is a linear system with $n=1$. Note that when $n=1$, SFC reduces to a linear form, which is equivalent to a linear mass-damper controller.
According to \textit{Property 1}, when there is a significant deviation in robot velocity state (e.g. ${x_2} > 1$), the spatial rate ratio of the nonlinear system (${\left. {{R_\text{sv}}} \right|_{{x_2} > 1,n > 1}}$) is greater than that of the linear system (${\left. {{R_\text{sv}}} \right|_{{x_2} > 1,n = 1}}$). This indicates that the velocity convergence of the nonlinear system under unit displacement is significantly higher than that of the linear system, i.e., the nonlinear system consumes more kinetic energy at high speeds. Therefore, when subjected to strong disturbances, the robot with nonlinear SFC can converge more quickly to stable regions. Conversely, when the velocity is lower than 1 (${x_2}\leq 1$), the nonlinear system converges more slowly than the linear system, indicating that robot with nonlinear SFC has a stronger ability to maintain its original motion at low velocities. This is physically manifested as being more compliant to drag forces. Figure~\ref{fig:phase_trajectory} displays the phase trajectories of the linear and nonlinear systems over a 10-second period, in line with the preceding analysis.

\vspace{2mm} \noindent \textbf{Property 2: (Stability).}
Assuming that the SFC has a constant input ${f_{\text{ext}}}$, there exists a unique equilibrium point at $\dot x = {\dot x^*}$, which is globally asymptotically stable.

\begin{proof}
For detailed analysis, see Appendix C
{\hfill $\blacksquare$\par}
\end{proof}

In a coupled state, robots and humans are closely physically contacted and generate continuous interactive forces between each other.
This force can be sustained and smooth, characterized by low frequency and even considered constant for a certain duration. The stability of the SFC ensures that, in this coupled state, the robot's movement will not diverge, enabling stable speed tracking and ensuring safety during the collaborative process.

Figure \ref{fig:SNNFC_P_n3} demonstrates that the SFC system will ultimately converge its velocity state to the singular line of $\dot x = {{\dot x}^ * }$, regardless of its $n$ value, when it deviates. This indicates spatial convergence of velocity. Additionally, at 10 seconds in Figure \ref{fig:SNNFC_P_n3}, the velocity converges to 0.2, indicating that the convergence of velocity for the nonlinear SFC system slows down at low speeds. In an actual mechanical system, the system will come to a stop when the velocity drops below the minimum speed threshold, preventing indefinite motion.

\vspace{2mm} \noindent \textbf{Property 3: (Passivity).}
If we express the SFC as an input-output equation with $f_\text{ext}$ as the input and $\dot x$ as the output, we can conclude that the SFC exhibits passive and dissipative behavior.
\begin{proof}
For detailed analysis, see Appendix C
{\hfill $\blacksquare$\par}
\end{proof}

SFC provides stable safety for the robot under constant external forces. However, when interacting with its environment, the robot may experience sudden disturbances or abrupt transitions from non-contact to contact. During this \textbf{impact phase}, external forces are often discontinuous or contain high-frequency components. The passivity of SFC ensures that the robot can effectively dissipate energy under any external force. This is achieved by preventing unrestricted energy accumulation during the control process, which in turn guarantees the reliability of the system.

\section{4. Frequency domain analysis}
Frequency domain analysis helps understand robot controller dynamic behavior, as well as guide controller design and tuning. It provides information on system response at different frequencies, such as determining the cutoff frequency for a balance between response speed and stability, and optimal gain for desired stability and performance at specific frequencies.

When dealing with nonlinear systems, frequency domain analysis typically involves the use of the \textit{describing function}~\cite{slotine1991applied}. However, for a SFC system that involve an n-th power feedback about the output, direct analysis using the describing function, which goes from input to output, is not feasible.
Since the SFC system is a bijection, the output uniquely determines the system's response. Thus, we suggest an \textit{inverse describing function} method to indirectly derive the describing function from the output. 
And the system parameters that are associated with the input are determined by utilizing the special relative relation at the cutoff frequency point. As a result, we have obtained Theorem 2 for the frequency response of the SFC system.

\vspace{2mm} \noindent
{\bf{Theorem 2: (SFC's describing function).}} 
Consider a system of the form (\ref{eq:with_mass}), with $m>0, \mu>0, n>0$. Then, the describing function of the SFC system, denoted as $N(B,\omega)$, can be obtained as
\begin{equation}\label{eq:N}
N(B,\omega ){\text{ }} \approx \frac{B}{{\sqrt {{{(mB\omega )}^2} + {{\left( {\mu {B^n}\Psi (n)} \right)}^2}} }}{e^{ - j\arctan \left( {\frac{{mB\omega }}{{\mu {B^n}\Psi (n)}}} \right)}}
\end{equation}
where $B$ is the amplitude of the output sinusoidal signals, $\omega$ is the frequency of the output signal, $\Psi \left( n \right) = \frac{{2\sqrt \pi  \Gamma \left( {1 + \frac{n}{2}} \right)}}{{\Gamma \left( {\frac{{3 + n}}{2}} \right)}}$ is a function related to the nonlinearity order, and $\Gamma \left( x \right) = \int_0^{ + \infty } {{t^{x - 1}}} {e^{ - t}}dt\left( {x > 0} \right)$ denotes the Gamma function, which is also known as Euler's second integral. 

The approximation in (\ref{eq:N}) provides a simplified expression for the describing function. Furthermore, the amplitude frequency response of the SFC system can be expressed as
\begin{equation}
|N(B,\omega)| \approx \frac{1}{\sqrt{(m\omega)^2+\left(\mu B^{n-1} \Psi(n)\right)^2}}
\end{equation}
and the phase frequency response can be expressed as
\begin{equation}
\angle N(B,\omega) \approx -\arctan\left(\frac{m\omega}{\mu B^{n-1} \Psi(n)}\right).
\end{equation}

\begin{proof}
For detailed analysis, see Appendix D
{\hfill $\blacksquare$\par}
\end{proof} 

Building upon the description of the function, we can further analyze the implications of key parameters in SFC design, which can provide guidelines for controller design. Consider a system of the form (\ref{eq:with_mass}), assuming $M>0$, $\mu>0$, $n>0$, then the following corollaries can be derived. 
\vspace{2mm} \noindent
{\bf{Corollary 1: (System bandwidth).}} The relationship between the input signal amplitude $A$ and the control bandwidth ${\omega_\text{c}}$ is given by 
\begin{equation} \label{eq:wc_AA}
{\omega_\text{c}} \approx \frac{{{{\left( {\mu \Psi \left( n \right)} \right)}^{\frac{1}{n}}}}}{m}{\left( {\frac{{{A}}}{{\sqrt 2 }}} \right)^{\frac{{n - 1}}{n}}}
\end{equation}

\begin{proof}
For detailed analysis, see Appendix D
{\hfill $\blacksquare$\par}
\end{proof} 

Analyzing the system bandwidth of a robot controller helps evaluate its response speed and improve robot control performance. The system bandwidth reflects the maximum frequency signal the controller can process, which is crucial for rapid response in robot applications. 
According to Equation (\ref{eq:wc_AA}), the system bandwidth of the robot under nonlinear SFC ($n>1$) is positively correlated with the amplitude $A$ of external force input. As the amplitude of external force decreases, the bandwidth of SFC becomes narrower, indicating that the robot will not over-respond to low-frequency and small-amplitude vibrations of the human body under SFC conditions. When $n=1$, the controller degenerates into the form of a linear time-invariant admittance control, and the system bandwidth remains constant and is independent of external force input.

\vspace{2mm} \noindent
{\bf{Corollary 2: (System time constant).}} 
The system time constant $\tau$ can be expressed as a function of $A$ as:
\begin{equation} \label{eq:TimeConstant1}
\tau  \approx \frac{m}{{{{\left( {\mu \Psi \left( n \right)} \right)}^{\frac{1}{n}}}{A^{\frac{{n - 1}}{n}}}}}
\end{equation}
\begin{proof}
For detailed analysis, see Appendix D
{\hfill $\blacksquare$\par}
\end{proof} 

The time constant represents the time required for the controller response to reach its peak value. In robot applications, the choice of time constant is crucial for ensuring that the controller can respond quickly and accurately to external disturbances, and achieve high-precision motion. As revealed by Equation (\ref{eq:TimeConstant1}), the time constant of SFC decreases with increasing input, indicating that a slight increase in force is sufficient for the collaborator to quickly modify the motion of the robot. And the time constant is determined by the variables $\mu$, $M$, and $n$, which can be adjusted by selecting appropriate values for these parameters in the equation.

\vspace{2mm} \noindent
{\bf{Corollary 3: (Correlation between 
variation of system gain and power-law).}} 
The correlation between variation of system gain ($\Delta {Q_w}$) and power-law ($n$) can be expressed as follows
\begin{equation}
    \Delta {Q_w} = 20w\frac{{1 - n}}{n}
\end{equation}
where, $Q_w$ represents the change in system gain caused by an input variation that is increased by a factor of ${10^w}$. 
\begin{proof}
For detailed analysis, see Appendix D
{\hfill $\blacksquare$\par}
\end{proof} 

The system gain change of the SFC nonlinear system ($n>1$) is negative, indicating that as the input signal amplitude increases, the attenuation effect becomes more pronounced. Moreover, the variation of system gain $\Delta {Q_w}$ is solely dependent on the power $n$, and there exists a lower limit. Specifically,
\begin{equation} \label{eq:lim1}
\mathop {\lim }\limits_{n \to \infty } \Delta {Q_w} = - 20w
\end{equation}

An excessively high power-law parameter in the controller could potentially cause it to become ill-conditioned (See Appendix G for details). By analyzing Equation (\ref{eq:lim1}), one can deduce that the system's attenuation does not continuously decrease as the SFC increases by $n$. Moreover, it's important to note that selecting a value for $n$ does not necessarily have to be a large number, but rather just adequate enough to fulfill the necessary requirements.

\section{5. Discretization constraint and parameter tuning}

When we apply SFC to real-world robot control, we need to take into account the problems of discretization constraint and parameter tuning. Discretization constraint refers to the fact that the controller operates in discrete time steps rather than continuous time, which may introduce non-passivity and instability. Parameter tuning is the process of finding the optimal values of the controller parameters that achieve the desired performance.

\subsection{5.1 Discretization constraint}

To regulate a robot's interaction with its environment, SFC relies on continuous data collection and real-time processing of sensor and actuator data, often supported by digital hardware and software. Consequently, a discrete-time controller is required. The SFC's discretized implementation can be expressed through the following set of equations.
\begin{equation} \label{eq:Discrete}
\left\{ \begin{gathered}
  \ddot x\left( t \right) = {m^{ - 1}}\left( {{f_\text{ext}}\left( t \right) - \mu {{\left| {\dot x\left( {t - 1} \right)} \right|}^{n - 1}}\dot x\left( {t - 1} \right)} \right) \hfill \\
  \dot x\left( t \right) = \dot x\left( {t - 1} \right) + \ddot x\left( t \right)\Delta T \hfill \\
  \dot {\bar x}\left( t \right) = g\dot x\left( t \right) \hfill \\ 
\end{gathered}  \right.
\end{equation}
where $\Delta T$ is the system sample time.

Equation (\ref{eq:Discrete}) indicates that SFC is attained through Euler integrals. However, these integrals can generate excess energy, which can compromise the energy dissipation characteristics of virtual dynamics, causing system vibration \cite{ferraguti2019variable,de2017passive}. We observed that the instability caused by Euler integrals is influenced by external force input, controller parameters, and system sampling time. In light of this, we propose discrete constraints that guarantee system stability once satisfied.

\vspace{2mm} \noindent
{\bf{Theorem 3: (Discrete control constraints).}} Consider a discretized SFC system of the form (\ref{eq:Discrete}) with parameters $n>0$, $m>0$, and $\mu>0$. Let ${f_\text{ext,max}}$ be the maximum value of external input force. Then, the upper limit of the sample time is constrained by

\begin{equation} \label{eq:constraint}
    \Delta T < 2m{\mu ^{\frac{{ - 1}}{n}}}{n^{ - 1}}{\left| {{f_\text{ext,max}}} \right|^{\frac{{1 - n}}{n}}}
\end{equation}

Furthermore, the relationship between the system bandwidth (${\omega_\text{c}}$) and the sample time under the discretized SFC is constrained by

\begin{equation} \label{eq:wwc1}
{\omega_\text{c}} < {2^{\frac{{n + 1}}{{2n}}}}\frac{{\Psi {{\left( n \right)}^{\frac{1}{n}}}}}{{\Delta Tn}}{\left| {\frac{{{f_\text{ext}}}}{{{f_\text{ext,max }}}}} \right|^{\frac{{n - 1}}{n}}}
\end{equation}

\begin{proof}
For detailed analysis, see Appendix E
{\hfill $\blacksquare$\par}
\end{proof} 

The upper bound for the system's longest sample time, ${\Delta T}$, has been established, along with an estimate of the system's bandwidth limit. By designing the system parameters based on these constraints, we can prevent instability resulting from excessively long sample time and tailor the sample time to meet the system's bandwidth requirements.

Due to the tightly coupled nature of the traction process, we analyzed the coupled stability of the system based on the practical model (See Appendix H) and obtained the following conditions for coupled stability.

\vspace{2mm} \noindent
{\bf{Theorem 4: (Coupled stability conditions).}} If the pHRI system, controlled by an SFC controller (\ref{eq:Discrete}) as shown in Figure \ref{fig:real_transform}, meets the following conditions, it can be regarded as having coupled stability.
\begin{equation}
    0 < Q < 1 \cap  0 < \Delta T\omega  < \pi
\end{equation}
where, $Q = \mu {B^{n - 1}}\Psi \left( n \right){m^{ - 1}}\Delta T$, $\omega$ represents the frequency of interactive force and $\Delta T$ denotes the system control cycle.
\begin{proof}
For detailed analysis, see Appendix I
{\hfill $\blacksquare$\par}
\end{proof}

\subsection{5.2 The SFC parameter tuning method}

In this section, we propose a parameter tuning method for the SFC in interactive applications, based on the corollaries discussed earlier. In environments where external force impact interference is present, robots need to meet two fundamental task prerequisites: withstanding external impact interference and complying with human traction force while exhibiting smooth and flexible movements. We outline the purpose and design rationale of each step in the Algorithm \ref{alg:auto}. The inputs to the algorithm consist of dynamic responsiveness requirements in human-robot interaction, which are presented in Table \ref{tab:require}. These requirements also correspond to two phase of resisting impact and complying with traction forces, which can occur simultaneously.

\begin{table}[t]
\small\sf\centering
\caption{The requirements for dynamic responsiveness in pHRI\label{tab:require}}
\begin{tabular}{ll}
\toprule
Notation&Explanation\\
\midrule
\texttt{$f_{\mathrm{ease}}$}& Maximum amplitude of external force \\& in the traction phase\\
\texttt{${{f_{{\mathrm{interf}}}}}$}& Maximum amplitude of external force\\& in the impact phase\\
\texttt{${{{\dot {\bar x}}_\mathrm{d}}}$}&Expected robot velocity in the traction phase\\
\texttt{${{\dot {\bar x}}_\mathrm{c}}$(${{\dot {\bar x}}_\mathrm{c}} \ge {{\dot {\bar x}}_\mathrm{d}}$)}&Acceptable robot velocity in the impact phase\\
\texttt{${\omega _\mathrm{c\_ease}}$}&System bandwidth in the traction phase\\
\texttt{$\Delta T$}& Sample time\\
\bottomrule
\end{tabular}
\end{table}

For the \textbf{impact force} described in Definition 2, the requirements are ${\varepsilon ^ + } = {\omega _{{\text{c\_max}}}}$ and ${f_\text{th}^ + }={f_{{\text{interf}}}}$:
\begin{equation} \label{eq:Force_Set2}
\small
\begin{gathered}
  {\mathcal{M}_{{\text{impact}}}} = \left\{ {\left. {{f_{{\text{ext}}}}\left( t \right)} \right|} \right. \hfill
{\omega _{{\text{ext,max}}}} \in \left( {{\omega _{{\text{c\_max}}}},\infty } \right)
\cup \left. {{f_{{\text{interf}}}} < {f_{{\rm{ext,max}}}}}  \right\}\hfill \\ 
\end{gathered}
\end{equation}

For the \textbf{traction force} described in Definition 3, the requirements are ${\varepsilon ^ - } = {\omega _{{\text{c\_ease}}}}$ and ${f_\text{th}^ - }={f_{{\text{ease}}}}$ :
\begin{equation} 
\small
\begin{gathered}
  {\mathcal{M}_{{\text{traction}}}} = \left\{ {\left. {{f_{{\text{ext}}}}\left( t \right)} \right|} \right. \hfill
{\omega _{{\text{ext,max}}}} \in \left( {0,{\omega _{{\text{c\_ease}}}}}  \right) \cap \left. {{f_{{\text{ease}}}} > {f_{{\rm{ext,max}}}}}  \right\}\hfill \\ 
\end{gathered}
\end{equation}

For the contactless situation, as defined in Definition 1, remains consistent across all tasks and does not require adjustments.

The algorithm includes five steps. First, it computes the power exponent $n$ that satisfies the given requirements, based on the correlation between force and velocity at different frequencies and amplitudes. Second, it determines the system bandwidth $\omega_{\text{c}\_\max}$ of the SFC under impact interference, while ensuring it meets the requirement for smooth traction functionality. Equations are provided to explain the calculations for $n$, $\omega_{\text{c}\_\max}$, and $\omega_\text{c\_ease}$, based on the known parameters and the desired force and velocity constraints. Third, it evaluates the system bandwidth to determine if it meets the constraint requirements, and make necessary modifications to address any deviations from the constraints. Fourth, it selects the virtual inertia of the system and calculate the apparent damping that meets the bandwidth requirements. Fifth, it calculates the coefficient for adjusting the gain of the system. Finally, it obtains controller parameters, $\mu, n, g$, that ultimately fulfill the requirements for interactivity. See Appendix F in the detailed derivation process.

Overall, the algorithm outlined in this section aims to automatically determine the appropriate values of controller parameters based on dynamic responsiveness requirements for human-robot interaction, and takes into account the force requirements for compliant movement and impact force limitation, as well as the system bandwidth of the SFC.

\begin{algorithm}[t]
\caption{Parameter auto-tuning}\label{alg:auto}
\begin{algorithmic}[1]
\Require ${f_{{\text{ease}}}},{f_{{\text{interf}}}},{{\dot {\bar x}}_\text{d}},{{\dot {\bar x}}_\text{c}},{\omega _\text{c\_ease}}, \Delta T, m$
\State $n \leftarrow {{\log \left| {\frac{{{f_{{\text{interf}}}}}}{{{f_{{\text{ease}}}}}}} \right|} \mathord{\left/
 {\vphantom {{\log \left| {\frac{{{f_{{\text{interf}}}}}}{{{f_{{\text{ease}}}}}}} \right|} {\log \left| {\frac{{{{\dot {\bar x}}_\text{c}}}}{{{{\dot {\bar x}}_\text{d}}}}} \right|}}} \right.
 \kern-\nulldelimiterspace} {\log \left| {\frac{{{{\dot {\bar x}}_\text{c}}}}{{{{\dot {\bar x}}_\text{d}}}}} \right|}}$
\State ${\omega _{\text{c}\_\max}} \leftarrow {\omega _\text{c\_ease}}{\left( {\frac{{{f_{{\text{interf}}}}}}{{{f_{{\text{ease}}}}}}} \right)^{\frac{{n - 1}}{n}}}$
\If{${\omega _{\text{c}\_\max}} > \frac{1}{{\Delta Tn}}{2^{\frac{{1 + n}}{{2n}}}}$}
    \State ${\omega _\text{c\_ease}} \gets \frac{2}{{\Delta Tn}}{\left( {\frac{{{f_{{\text{ease}}}}}}{{\sqrt 2 {f_{{\text{interf}}}}}}} \right)^{\frac{{n - 1}}{n}}}$
\EndIf
\State $\mu \gets \frac{{{{\left( {m{\omega _\text{c\_ease}}} \right)}^n}}}{{\Psi \left( n \right)}}{\left( {\frac{{\sqrt 2 }}{{{f_\text{ease}}}}} \right)^{n - 1}}$
\State $g \leftarrow {{\dot {\bar x}}_\text{d}}{\left( {\frac{\mu }{{{f_{{\text{ease}}}}}}} \right)^{\frac{1}{n}}}$
\Ensure $\mu, n, g$
\end{algorithmic}
\end{algorithm}

\section{6. Simulations}

The following section presents a comparative analysis of the dynamic properties associated with SFC, Linear Admittance Control (L-AC), and Nonlinear Admittance Control (N-AC) in both the time and frequency domains. It explains the reasons why SFC can achieve traction compliance and impact resistance. Furthermore, it discusses the performance differences resulting from constructing the nonlinear damping term through different approaches using SFC and N-AC. Corollaries derived from SFC, as presented in Section 5, are compared with numerical iterative solutions to ensure model accuracy and theoretical validity. Additionally, the parameter constraints imposed by discretization under SFC, as discussed in Section 6, are confirmed through comparison. All simulations are performed within the Simulink environment.

The L-AC (\ref{eq:AD}) utilized for controller comparison is presented as follows
\begin{equation} \label{eq:AD}
  \left\{ \begin{gathered}
 {m_{\mathrm{A}}}\ddot x + {\mu _{\mathrm{A}}}\dot x = {f_{\mathrm{ext}}} \hfill \\
 {\dot { \bar x}} = {g_{\mathrm{A}}}\dot x \hfill \\
\end{gathered}  \right.
\end{equation}
where, ${m_{\mathrm{A}}>0}$, ${\mu_{\mathrm{A}}>0}$, and ${g_{\mathrm{A}}>0}$ denote the virtual inertia, virtual damping, and gain parameters of L-AC, respectively.

Since there is a very limited availability of nonlinear admittance controllers in the literature, and their control objectives are not aligned with our research, we selected a study by \cite{lai2014improving} that was formally closest to our requirements. We then made minor adjustments while ensuring the stability conditions of the controller. The modified N-AC, used for controller comparison, is described as follows
\begin{equation}  \label{eq:N-AC}
\left\{ \begin{array}{l}
{m_{\rm{N}}}\ddot x + \left( {{\mu_\mathrm{N}} + {\alpha_\mathrm{N}}\left( {1 - {e^{ - \frac{{{f_{{\rm{ext}}}}^2}}{{{\sigma_\mathrm{N}}^2}}}}} \right)} \right)\dot x = {f_{{\rm{ext}}}}\\
  {\dot { \bar x}} = {g_\mathrm{N}}\dot x \hfill \\
\end{array} \right.
\end{equation}
where, ${m_\mathrm{N}}>0$, ${\mu_\mathrm{N}}>0$, and ${g_\mathrm{N}}>0$ denote the virtual inertia, virtual damping, and gain parameters of N-AC, respectively. $\sigma_\mathrm{N}>0$ determines the point at which nonlinear damping becomes significant, while ${\alpha_\mathrm{N}>0}$ represents the additional damping added to the system when ${f_{{\rm{ext}}}}>>{\sigma _\mathrm{N}}$.

Originally, this controller aimed to improve the transient performance of contact/machining forces between the robot and the workpiece. When the contact force ${f_{{\rm{real}}}}$ approaches the desired value ${f_{{\rm{desire}}}}$ (i.e. ${f_{{\rm{ext}}}} = {f_{{\rm{real}}}} - {f_{{\rm{desire}}}} = 0$ ), the controller can achieve higher damping to accelerate the convergence of the contact force and reduce overshooting. Although its control objectives are contrary to ours, we can achieve similar control objectives by making slight adjustments, such as increasing the nonlinear damping as the force increases, as described in (\ref{eq:N-AC}).

\subsection{6.1 Frequency domain response}

\begin{figure*}[t]
\centering
\subfloat[]{\includegraphics [scale=1.1]{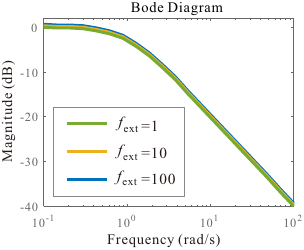}%
\label{fig:SNNFC_M1_u1_n1}}
\subfloat[]{\includegraphics [scale=1.1]{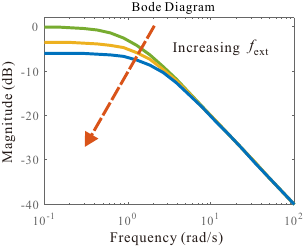}%
\label{fig:Bode_NAC}}
\subfloat[]{\includegraphics [scale=1.1]{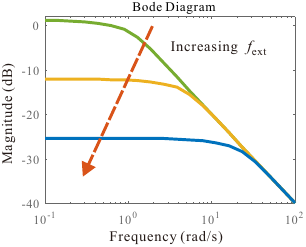}%
\label{fig:SNNFC_M1_u1_n3}}
\caption{Amplitude response in frequency domain. (a) L-AC (${m_{\rm{A}}} = {\mu _{\rm{A}}} = G_{\rm{A}}= 1$) (b) N-AC(${m_{\rm{N}}} = {\mu _{\rm{N}}} = g_{\rm{N}}={\alpha_{\rm{N}}}= 1,{\sigma _{\rm{N}}}=10$). (c) SFC ($n=3> 1$, ${m} = {\mu} = g = 1$). The dashed arrow indicates an increase in external force.}
\label{fig:SNNFC_response}
\end{figure*}

The numerical simulation conducted through Simulink is instrumental in verifying the frequency-domain response of SFC. The system input signal is represented by ${f_{{\rm{ext}}}} = A\sin \left( {\omega t} \right)$, where input amplitude $A$ takes the values of $\left\{ {1, 10, 100} \right\}$ and the frequency $\omega$ varies from 0~Hz to 100~Hz. Figure~\ref{fig:SNNFC_response} presents the comparative analysis between L-AC, N-AC and the proposed SFC. It should be noted that bode diagrams can be translated up and down as a whole by adding a gain at the output, thus rendering absolute magnitude insignificant.  Instead, emphasis should be directed towards the relative relationship of gain values in the Bode plot under varying input amplitudes.

Figure~\ref{fig:SNNFC_M1_u1_n3} demonstrates that when the power $n>1$, the system transitions from amplification to attenuation with increasing input amplitude $A$. Furthermore, as $A$ increases, the system's attenuation becomes more pronounced. The observed \textit{stratification phenomenon} is dependent on the input amplitude $A$, resulting in a monotonic shift of the system's frequency domain response curve as $A$ increases. This response behavior is primarily dictated by the term ${A^{\frac{1}{n}-1}}$ within (\ref{eq:N_A0}). As illustrated in Figure~\ref{fig:Bode_NAC}, another nonlinear controller, N-AC, exhibits a similar stratification phenomenon. The nonlinearity in N-AC, described by the restricted exponential function in equation (\ref{eq:N-AC}), ensures that the nonlinear damping does not continuously increase with an increase in external force. Consequently, the attenuation in N-AC is finite and not as pronounced as in SFC. In clear contrast, the L-AC system ($n=1$) exhibits no stratification and independence of $A$ in its response (as depicted in Figure.~\ref{fig:SNNFC_M1_u1_n1}), analogous to Newtonian fluid properties.

\subsection{6.2 Time domain response}

Step response analyses are performed on systems (\ref{eq:with_mass}), (\ref{eq:N-AC}) and (\ref{eq:AD}), which represent the three controllers under consideration. The input signals ${f_{{\text{ext}}}} = \left\{ {0.5,5,50} \right\}$~N are used to test the performance of both controllers. Time constants and steady-state values for each controller are determined and subsequently compared. The controller parameters correspond to the fixed manipulator parameters in Table~\ref{tab:ParaS}, and a detailed explanation of the selection process can be found in Section 7.1.

\begin{table}[t]
  \small\sf\centering
  \caption{Parameters Setting}
  \resizebox{0.49\textwidth}{30mm}{
  \begin{tabular}{ccccc}
    \toprule
   Controller Type & L-AC & SFC & N-AC \\
    \midrule
   Controller Parameters  & & Fixed Manipulator \\
    \midrule
    Power Law & ${n_\mathrm{A}} = 1$ & ${n}  = 3 $ & - \\
    Scaled Mass & ${m_\mathrm{A}} = 1$ & $ {m} = 1$ & ${m_\mathrm{N}} = 1$ \\
    Scaled damping & ${\mu_\mathrm{A}} = 17$ & ${\mu} = 393$ & ${\mu_\mathrm{N}} = 15.5$ \\
    Scaling Factor & ${g_\mathrm{A}} = 0.17$ & ${g} = 0.21$ & ${g_\mathrm{N}} = 0.17$ \\
    Additional Damping & - & - & ${\alpha_\mathrm{N}}= 25$ \\
    Nonlinear Point & - & - & ${\sigma_\mathrm{N}}=20$ \\
    Sample Time (s) & $\Delta T =   0.002$ & $\Delta T =   0.002$ & $\Delta T =   0.002$ \\
    \midrule
     & & Mobile Manipulator \\
    \midrule
    Power Law & ${n_\mathrm{A}} = 1$ & ${n}  = 3 $ & - \\
    Scaled Mass & ${m_\mathrm{A}} = 1$ & $ {m} = 1$ & ${m_\mathrm{N}} = 1$ \\
    Scaled damping & ${\mu_\mathrm{A}} = 8$ & ${\mu} =20$ & ${\mu_\mathrm{N}} = 7$ \\
    Scaling Factor & ${g_\mathrm{A}} = 0.045$ & ${g} = 0.04$ & ${g_\mathrm{N}} = 0.45$ \\
    Additional Damping & - & - & ${\alpha_\mathrm{N}}= 7$ \\
    Nonlinear Point & - & - & ${\sigma_\mathrm{N}}=10$ \\
    Sample Time (s) & $\Delta T =   0.02$ & $\Delta T =   0.02$ &  $\Delta T =   0.02$ \\
    \bottomrule
  \end{tabular}
  }
  \label{tab:ParaS}
\end{table}

\begin{figure*}[t]
\centering
\subfloat[]{\includegraphics [scale=1]{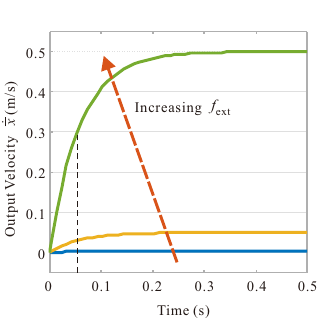}%
\label{fig:Time_n1}}
\hfil
\subfloat[]{\includegraphics [scale=1]{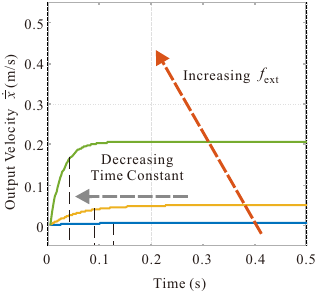}%
\label{fig:Step_NAC}}
\hfil
\subfloat[]{\includegraphics [scale=1]{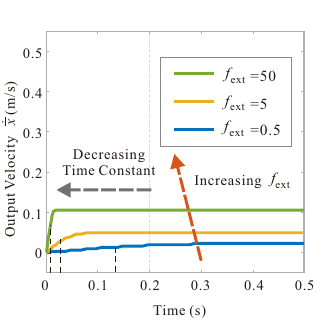}%
\label{fig:Time_n3}}
\caption{Step response in time domain. The red dashed arrow is the direction of the external force increase. The direction of the grey dashed arrow represents the decrease in the time constant. (a) L-AC ($n = 1$) (b) N-AC (c) SFC ($n=3>1$)}
\label{fig:Time_response}
\end{figure*}
\subsubsection{6.2.1 Time constant.}
All three controllers have reached a steady state, as depicted in Figure \ref{fig:Time_response}. While L-AC presents an unchanged time constant regardless of the input signal (Figure.~\ref{fig:Time_n1}), SFC demonstrates a distinct behavior in this regard. Specifically, SFC's time constant gradually decreases with the increase of input force (as shown in Figure.~\ref{fig:Time_n3}). Such a phenomenon suggests that SFC can quickly converge when faced with high external forces, thereby enabling efficient collaborator adjustments. On the other hand, small-amplitude perturbations result in substantial delays, allowing the collaborator to operate with reduced force and prevent unwanted vibrations arising from slight tremors in human motion.
N-AC exhibits similar characteristics to SFC, but due to the gradual convergence of its nonlinear damping term with increasing external forces (as shown in Figure.~\ref{fig:Step_NAC}), its nonlinearity is limited, and the changes in the time constant are not as significant as those in SFC.

\subsubsection{6.2.2 Steady-state value.}

When the input step signal is 5 N, the three controllers yield identical steady-state values (0.05 m/s), as depicted by the yellow curve in Figure \ref{fig:Time_response}. This indicates that they will generate the same responsive velocity under a dragging force of 5 N. However, when subjected to small forces (blue curve in Figure~\ref{fig:Time_response}), the velocity response generated by SFC surpasses that of L-AC, suggesting a higher likelihood of the robot conforming to traction intentions under the former's control. In contrast, during large impact forces (green curve in Figure~\ref{fig:Time_response}), L-AC results in motion speeds two times greater than those exhibited by SFC, thereby increasing safety hazards for close-contact collaborations. Therefore, SFC limits the speed to a reasonable level, promoting safe human-robot collaboration. N-AC exhibits a response similar to SFC, but due to the limited non-linear damping, its attenuation of large impact forces (green curve in Figure~\ref{fig:Step_NAC}) is not as pronounced as SFC.

\subsubsection{6.2.3 Impulse response analysis}
To test the controller response under the combined effect of traction and impact force, we applied the following external force input (\ref{eq:impulse}). We started by applying a constant traction force of 5 N, and then at 0.6 s, we simulated the impact force input by applying a 50 N pulse for a duration of 0.4 s.

\begin{figure}[t]
\centering
\includegraphics [scale=1.1]{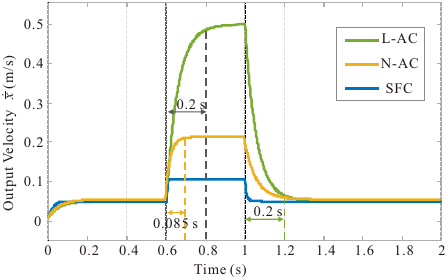}%
\caption{Impulse response in time domain.}
\label{fig:Impulse_all}
\end{figure}

\begin{equation} \label{eq:impulse}
{f_\text{ext}}\left( t \right) = \left\{ {\begin{array}{*{20}{c}}
{5\ {\rm{N}}}&{t{\rm{ < 0}}{\rm{.6\ s}}}\\
{50\ {\rm{N}}}&{{\rm{0}}{\rm{.6\ s}} \le t {\rm{ < }}1\ {\rm{s}}}\\
{5\ {\rm{N}}}&{1\ {\rm{s}} \le t }
\end{array}} \right.
\end{equation}

The response of the three controllers is depicted in the Figure \ref{fig:Impulse_all}. At a constant traction force of 5 N, all three controllers exhibit the same output velocity of 0.05 m/s, indicating consistent response amplitudes to the traction force.
However, when a 50 N impact force occurs, the linear controller (L-AC) experiences a sudden jump in output velocity to 0.5 m/s. In contrast, the output velocity jumps of the nonlinear controller N-AC and SFC are smaller than that of L-AC, as their system gains attenuate with increasing input amplitudes. Additionally, due to the upper limit ${\mu _{\rm{N}}} + {\alpha _{\rm{N}}}$ of nonlinear damping in N-AC, its ability to suppress velocity jumps is lower than that of SFC.

In addition, due to the explicit correlation between the nonlinear damping term $\left( {{\mu _{\rm{N}}} + {\alpha _{\rm{N}}}\left( {1 - {e^{ - \frac{{{f_{{\rm{ext}}}}^2}}{{{\sigma _{\rm{N}}}^2}}}}} \right)} \right)$ of N-AC and the external force (${f_{{\rm{ext}}}}$), the nonlinear damping term of N-AC significantly increases at 0.6 s when a 50 N impact force occurs, reaching steady-state within 0.085 s. However, when the 50 N impact force is removed, the nonlinear damping term of N-AC decreases significantly, resulting in a time delay for it to reach steady-state (0.2 s). In contrast, the nonlinear damping of SFC is only dependent on the velocity state of the system itself. Thus, even after the impact force disappears, SFC can maintain a high level of nonlinear damping based on the current velocity, enabling it to quickly return to a stable state.

\subsection{6.3 Analysis verification of SFC's dynamic properties}

Besides the comparison between SFC, L-AC and N-AC, a numerical verification of the analytical results regarding SFC corollaries is also performed. The accuracy of the analytical solutions for SFC's dynamic properties is evaluated against numerical simulation results to validate their correctness.
\subsubsection{6.3.1 System bandwidth.}

In Figure~\ref{fig:SNNFC_M1_u1_n3}, the numerical solution for the SFC system bandwidth corresponds to the -3 dB point of each curve. The resulting analytical solution (\ref{eq:wc_A}) is compared against the numerical solution, with the comparison results found in Table~\ref{tab:bandwidth}. The relative error between the analytical and simulated solutions is less than 2\%, indicating the validity of the analytical approach. We think that the error in the analytical solution arises from the use of the Describing Function method for analyzing nonlinear controllers (\cite{slotine1991applied}). This method only considers the fundamental frequency response and assumes that higher-order harmonics have lower energy and are filtered out by the linear part of the system. However, in reality, we believe that some residual effects of higher-order harmonics still exist, leading to a 2\% error.

\begin{table}[t]
  \small\sf\centering
  \caption{Comparison of analytical and numerical solutions\\ of SFC's system bandwidth}
    \begin{tabular}{cccc}
    \toprule
    Force   &  Analytical   & Numerical & Relative  \\
    Amplitude (N)  &  ${\omega_\text{c}}\left( {{\text{Hz}}} \right)$&  ${\omega_\text{c}}\left( {{\text{Hz}}} \right)$ & Error \\
    \midrule
    1   & 1.05  & 1.04  & 1\% \\
    10     & 4.90  & 4.90  & 0\% \\
    100     & 22.75  & 23.20  & 2\% \\
   
    \bottomrule
    \end{tabular}%
  \label{tab:bandwidth}%
\end{table}%
 
\begin{table}[t]
  \small\sf\centering
  \caption{Comparison of analytical and numerical solutions\\ of SFC's time constant}
    \begin{tabular}{cccc}
    \toprule
    Force   &  Analytical   & Numerical & Relative  \\
    Amplitude (N)  &  ${\tau}\left( {{\text{s}}} \right)$&  ${\tau}\left( {{\text{s}}} \right)$ & Error \\
    \midrule
    0.5   & 0.162  & 0.145  & 10\% \\
    5     & 0.035  & 0.032  & 9\% \\
    50     &0.0076  & 0.007  & 8\% \\
  
    \bottomrule
    \end{tabular}%
  \label{tab:timeconstant}%
\end{table}%

\begin{table}[t]
  \small\sf\centering
  \caption{Comparison of analytical and numerical solutions\\ of SFC's System Gain Variation}
    \begin{tabular}{cccc}
    \toprule
    Power-law &  Analytical   & Numerical & Relative  \\
    $n$ &   $\Delta {Q_2}$ (dB) &   $\Delta {Q_2}$ (dB) & Error \\
    \midrule
    1   & 0.00  & 0.00  & 0\% \\
    3     & -26.67  & -26.67  & 0\% \\
    10     & -36.00  & -36.00  & 0\% \\
    100     &-39.60  & -39.60  & 0\% \\
  
    \bottomrule
    \end{tabular}%
  \label{tab:gain}%
\end{table}%
\subsubsection{6.3.2 Time constant.}
The time constant measures the duration for the system's response to attain 63\% of its steady-state value following a step input (from initial zero conditions). 
A numerical estimate of SFC's time constant corresponding to Figure~\ref{fig:Time_n3} and an analytical estimate given by (\ref{eq:TimeConstant1}) can be calculated, with a comparison between both approaches presented in Table~\ref{tab:timeconstant} under various inputs. The analytical and simulated solutions have a relative error of 10\%. For the same reason of approximating with Describing Functions, we attribute this error to the residual effects of higher-order harmonics. Unlike the approximation error in the system bandwidth, which affects the numerator and is relatively small, the approximation error in this case affects the denominator, resulting in a relative error of 10\%. Despite this limitation, this approximate estimation suffices for effective system design purposes.

\subsubsection{6.3.3 System gain variation.}

The analytical solution for the system gain variation is obtained through (\ref{eq:dQ}). Numerical simulations implementing the sine sweep test are conducted for various power-law values $n = \left\{ {1,3,10,100} \right\}$, as $A_0=1$, $m=2$, and ${m} = {\mu} = g = 1$ remain consistent. A comparison between the analytical and numerical solutions reveals a high degree of similarity, as represented in Table~\ref{tab:gain}. As stipulated by (\ref{eq:lim}), the lower limit of gain variation equates to -40 dB. For $n=3$, the resulting gain variation is equal to 66.7\% of the lower limit, enabling clear differentiation between different inputs. Moreover, when $n=100$, a value of -39.6 dB is achieved, thereby validating the existence of attenuation's lower limit. Accordingly, moderate rather than excessively large choices for $n$ are suitable since they suffice to produce reliable results without unnecessary complexity.

\subsection{6.4 SFC's discretization constraints}

\begin{figure}[t] 
    \centering
    \includegraphics[width=7.2cm]{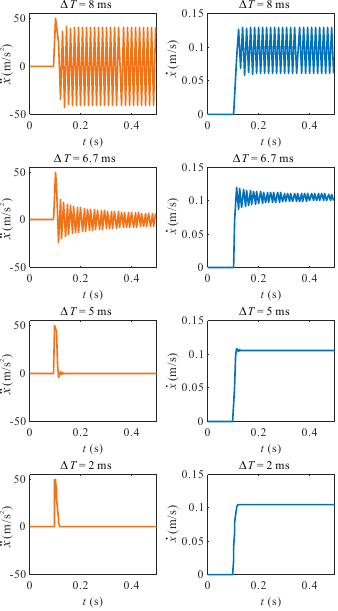}
    \caption{SFC's step responses at different sample times}
    \label{fig:constraint}
\end{figure}

To verify the discretization constraint (\ref{eq:constraint}), the step response of SFC is evaluated at different sample times. SFC parameters are obtained from the first row in Table~\ref{tab:ParaS}, and a step input of 50 N occurs at 0.1 s. Step responses are examined for various sample times ($\Delta T = \left\{ {8,6.7,5,2} \right\}$~ms) separately, as shown in Figure~\ref{fig:constraint}. The upper limit of SFC's sample time, according to (\ref{eq:constraint}), is $\Delta T=6.7$ ms.
As sample time exceeds this upper limit (e.g., $\Delta T=8>6.7$ ms), the controller generates acceleration oscillations, hampering speed convergence. As the upper limit is approached, the control output converges slowly with oscillation ($\Delta T=6.7$ ms). Once the upper limit is met ($\Delta T=$ 5 or 2~ms), oscillations cease to exist within the controller. Therefore, by referring to the constraint (\ref{eq:constraint}), SFC's sample time can be chosen based on the input force range to avoid oscillations stemming from system discretization.

\subsection{6.5 Discussion}

Based on the aforementioned simulation experiments, the following conclusions can be drawn:

1) Leveraging the introduction of a nonlinear power-law, SFC exhibits superior dynamic properties as compared to L-AC when handling impact disturbances in both time and frequency domains. SFC's bode diagram depicts a stratification phenomenon that enables it to react distinctively towards various input forces, thereby clarifying its ability to differentiate between traction and impact forces regarding human-robot collaboration. SFC's bandwidth becomes narrower as input declines, indicating that the robot is not overly responsive to low-frequency, small-amplitude vibrations of the human body under SFC. The time constant of SFC's step response diminishes as force input increases, signifying that slight raises in force suffice for collaborators to quickly modify robot motion. On the other hand, L-AC's bode diagrams with different inputs exhibit precise overlap, while its time constants for step response remain identical, amplifying the difficulties of resisting high-intensity impact.

2) Comparing SFC and N-AC, both controllers incorporate nonlinear damping. SFC's nonlinear damping term only depends on its velocity state variable, while N-AC explicitly includes the external force variable in its damping term. Both nonlinear controllers can achieve layered attenuation in frequency response with increasing external force input. However, as the external force increases, N-AC's nonlinear damping converges to a constant value, resulting in a less pronounced ability to suppress high-intensity external forces compared to SFC. Additionally, N-AC's nonlinear damping is directly related to the external force, so when the external impact force is removed, the damping immediately decreases, leading to a longer time for N-AC to return to steady state. In contrast, SFC's nonlinear term is only dependent on its own state, so it maintains higher damping when there is a significant velocity deviation, pulling it back to the steady-state value more quickly.

3) The analytical corollaries proposed in Section 5 concerning SFC's dynamic properties, including system bandwidth, time constant, and gain variation, are verified through comparison with numerical iterative solutions developed in Simulink. The relative error margin does not exceed 10\%, validating the feasibility of applying analytical solutions to guide SFC parameter settings. 

4) Robots with low control frequencies must consider the effect of discretization, as failure to do so may result in divergent robot acceleration. Section 6 presents the discrete control constraint for robots, which can be validated through testing robot responses at varying sample times. Taking into account the constraint ensures stable step responses for robots, thereby enabling effective utilization of SFC for robots with low control frequencies.

\section{7. Real world experiments}
The real-world experimental setup entailed a human and a robot holding opposite sides of a tray containing a glass of water. The robot, guided by force perception and compliant with the human, was able to transport the glass of water to any desired location.
The experimental task was constructed within the Cartesian space, whereby virtual dynamic controls were utilized to accomplish three-dimensional translational motion, resulting in the robot obediently adhering to the cooperator's applied traction. Three-dimensional rotation was sustained through a Proportional-Derivative (PD) control scheme, which facilitated horizontal orientation maintenance.

The study conducted a comparison of the control efficiency of L-AC (\ref{eq:AD}), N-AC (\ref{eq:N-AC}) and the proposed SFC (\ref{eq:with_mass}), both functioning as virtual dynamic control modules within the pHRI framework depicted in Figure \ref{fig:framework}. The discretization process for these three controllers follows the equation (\ref{eq:Discrete}).
For the robot platform in Figure \ref{fig:framework}, we employed a fixed manipulator and a mobile manipulator successively. 
The two robots have different Jacobian matrices for their differential inverse kinematics modules, namely $\textbf{J}_f(\textbf{q}(t)) \in {\mathbb{R}^{{6} \times {6}}}$ and $\textbf{J}_m(\textbf{q}(t)) \in {\mathbb{R}^{{6} \times {9}}}$. Additionally, they are operated at frequencies of 500Hz and 50Hz, respectively. The fixed manipulator used is the UR16e, which can achieve a sampling frequency of 500Hz. The mobile manipulator involves mounting the UR16e manipulator on a mecanum mobile platform. This mobile platform has a mass of 370kg. To ensure control stability under high inertia, the sampling frequency of the mobile platform is reduced to 50 Hz. The mobile manipulator treats the manipulator and the mobile platform as a whole, forming a 9-degree-of-freedom robot. To ensure synchronization in control, the manipulator and the mobile platform need to have the same sampling frequency. Hence, the overall control frequency of the mobile manipulator is 50 Hz. This is the reason for the big difference in sampling frequencies between the two systems.

For the fixed manipulator, we evaluated the dynamic performance of L-AC, N-AC, and SFC during the traction phase, impact phase, and simultaneous impact and traction phase. Additionally, for the mobile manipulator, we compared the L-AC, N-AC, and SFC under low system bandwidth conditions. An accompanying video showcasing the results of the experiments is included in Extension 1.


\subsection{7.1 Parameters setting}


\begin{figure} [t]
  \centering
  \includegraphics[scale=1]{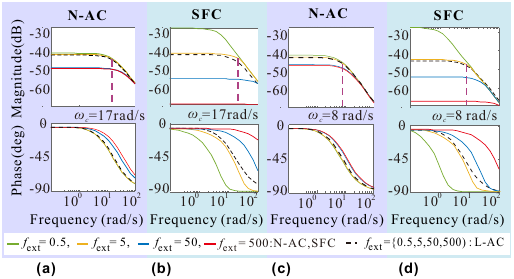}
  \caption{Comparison of Bode diagrams for L-AC, N-AC, and SFC in both fixed and mobile manipulators. The Bode diagram for L-AC is represented by the black dashed line, while the colored curves depict the Bode diagrams for N-AC and SFC. (Subplot a-b) is for the fixed manipulator, while (subplot c-d) is for the mobile manipulator.
  }
  \label{fig:Exp_bode2}
\end{figure}

\subsubsection{7.1.1 Parameters applied to fixed manipulator.} 

For human-robot collaboration based on force perception, it is important to ensure that the robot can be guided easily by the human collaborator. Previous research has shown that human daily activities performed on a force platform typically occur in the frequency range of 0.3-3.5 Hz (~\cite{khusainov2013real,plasqui2013daily}). Therefore, we chose a system bandwidth of ${\omega _{c}} = 2.7{\text{ Hz}}$ to filter out high frequency noise and ensure the comfort of the collaborator during dragging. Additionally, we expect the robot to achieve a compliance speed of 0.05~m/s when ${f_{{\text{ext}}}} = 5$ N to ensure the safety of traction. When an impact occurs, the disturbance force is typically no more than 60 N. The experimental setup used a UR16e fixed manipulator with a control frequency of 500~Hz.
The parameters of L-AC, N-AC and SFC for application on a fixed manipulator were selected based on the aforementioned requirements and parameter auto-tuning algorithm \ref{alg:auto}, and are presented in Table~\ref{tab:ParaS}. Simulink was employed to conduct a simulation of the three controllers under the selected parameters.

As shown in Figure~\ref{fig:Exp_bode2}, the black dashed line represents the frequency domain response of L-AC, while the colored solid lines represent the frequency domain responses of N-AC and SFC. 
While the response curves of L-AC overlap for input forces with different amplitudes, the response curves of N-AC and SFC exhibit distinct layered patterns.
At ${f_{{\text{ext}}}} = 5$~N, the curves of L-AC, N-AC and SFC coincide, indicating consistent responses near the optimal force required for human manipulation and suggesting that N-AC and SFC can achieve the same level of compliance effect as L-AC. 
However, when subjected to an increased external impact force, such as at ${f_{{\text{ext}}}} = 50$~N, the gain of SFC is significantly reduced, while N-AC experiences a limited attenuation in gain, and L-AC maintains a constant gain. This demonstrates that SFC has a superior ability to resist high-intensity forces compared to both N-AC and L-AC.

\begin{figure*}[th]
\centering
\includegraphics[width=6in]{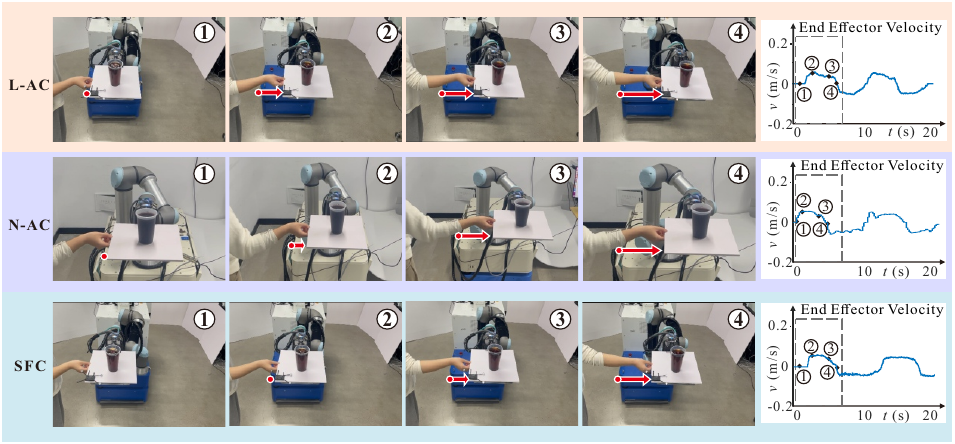}%
\caption{\textit{The traction phase} - experimental snapshots: initialization by human operator, followed by forward traction (The duration of the sequence is 5s). The numbers in the graph correspond to the numbers on the curve, representing the moment of the snapshot. The red arrow is the direction of the traction force.}
\label{fig_sim}
\end{figure*}

\begin{figure}
  \centering
  \includegraphics[width=3.2in]{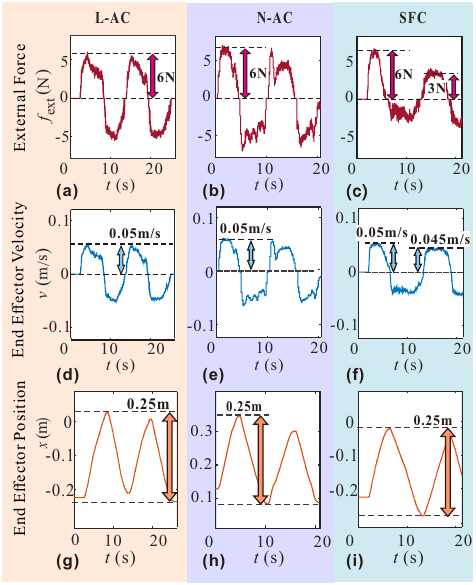}
  \caption{Results of the \textit{traction phase} on the fixed manipulator\\
  \hspace{1.1cm} column 1: (subplots a-d-g) L-AC - traction
  \\
  \hspace{1.1cm} column 2: (subplots b-e-h) N-AC - traction
  \\
  \hspace{1.1cm} column 3: (subplots c-f-i) SFC - traction}
  \label{fig:E_6N_3}
  \end{figure}

\begin{figure*}[t]
\centering
\includegraphics[width=6.5in]{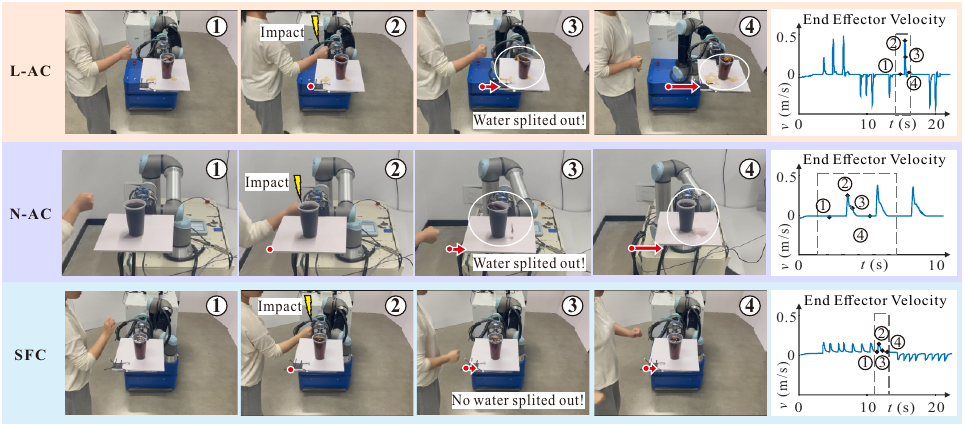}%
\caption{\textit{The impact phase} - experimental snapshots: Initialization by a human operator for impact response testing (The duration of the sequence is 0.5s). The impact force occurs at the moment of the snapshot in number 2. The red arrow shows the direction and distance of the end effector after the impact.}
\label{fig_sim2}
\end{figure*}

\subsubsection{7.1.2 Parameters applied to mobile manipulator.}
SFC is suitable not only for robots with high control frequencies but also for those with low control frequencies, as long as SFC parameters conform to the discretization constraint (\ref{eq:constraint}). 
To prevent jitter issues in a mobile manipulator, it is necessary to ensure that the control frequency between the mobile platform and manipulator is synchronized. 
For instance, if the control frequency of the mobile platform reaches 50~Hz, a rapid discrete sample time of no less than 0.02~s will be required.
As demonstrated by the lower half of Table \ref{tab:ParaS}, specific parameter redesigning is necessary for the mobile manipulator. Notably, during experiments where the maximal external impact force reached ${f_\text{ext,max }} = 70$~N, inequality (\ref{eq:constraint}) indicated that the sample time should be less than 0.023~s. The sample time of 0.02~s implemented in the mobile manipulator met this requirement and assured the stability of the system under discrete control.

Figure~\ref{fig:Exp_bode2} illustrates that under both mobile and fixed mechanical setups, L-AC, N-AC, and SFC exhibit similar bode diagram patterns, indicating their ability to achieve equivalent dynamic characteristics on both mobile and fixed machinery.
The system bandwidth's upper limit was 9.6~rad/s at around 10~N, given the sample time, as evidenced by inequality (\ref{eq:wwc1}). The controller bandwidth deployed for the mobile manipulator (8~rad/s) satisfied the constraint limit. However, the long sample time resulted in the mobile manipulator's bandwidth decreasing to 8~rad/s, compared to the fixed manipulator's 17~rad/s. Nevertheless, the system bandwidth could not be too low; otherwise, it would lead to a significant time delay and impact the human-robot interaction's comfort level. Hence, when the sample time is extended, there is a trade-off in parameter design for the system bandwidth.

\subsection{7.2 Comparison of results on the fixed manipulator}

\begin{figure*}
  \centering
  \includegraphics[width=6.5in]{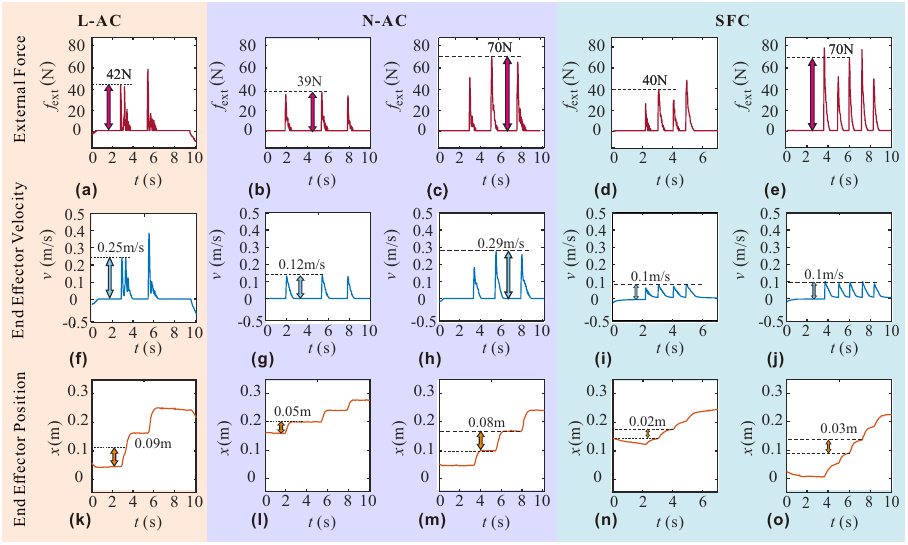}
  \caption{Results of the \textit{impact phase} on the fixed manipulator\\
  \hspace{1.1cm} column 1: (subplots a-f-k) L-AC Moderate impact
  \hspace{1.1cm} column 2: (subplots b-g-l) N-AC Moderate impact\\
  \hspace{1.1cm} column 3: (subplots c-h-m) N-AC Intense impact
  \hspace{1.2cm} column 4: (subplots d-i-n) SFC Moderate impact\\
  \hspace{1.1cm} column 5: (subplots e-j-o) SFC Intense impact}
  \label{fig:E_60N_3}
  \end{figure*}

The experiments performed on the fixed manipulator were categorized into three phases - \textit{the traction phase}, \textit{the impact phase}, and \textit{the simultaneous impact and traction phase}. This division allowed for a comparison of the response performance of L-AC, N-AC, and SFC under different external forces. The robot behaviors are shown in the accompanying video. 

\subsubsection{7.2.1 The traction phase. }

To compare the compliance ability, the robot's end-effector was dragged to move repetitively with a gentle force of around 6 N. In Figure \ref{fig_sim}, three snapshot sequences are shown depicting smooth dragging actions: the initial contact occurs at 2.5 s (snapshot 1) and within the following 5 s, the water cup is easily propelled forward (snapshots 2-4). The responses using L-AC, N-AC, and SFC were similar, resulting in an approximate velocity output of 0.05 m/s (Figures \ref{fig:E_6N_3}d, \ref{fig:E_6N_3}e, and \ref{fig:E_6N_3}f). Eventually, the end-effector achieved a reciprocating motion of approximately 0.25 m. These results indicate that within a force range facilitating human control, SFC, NAC, and L-AC can all achieve the same performance effect. Moreover, SFC amplifies small traction forces, which is exhibited in Figures \ref{fig:E_6N_3}c and \ref{fig:E_6N_3}f. Specifically, when the external force input amplitude equals 3 N at about 15 s, the system's velocity response rate is 0.045 m/s, nearly equivalent to the effect of 6 N external force dragging, suggesting that the collaborator can effortlessly drag the robot during cooperative motion under SFC.

\subsubsection{7.2.2 The impact phase. } 
To evaluate the anti-interference performance, the operator applied external forces of 40 N and 60 N to the end-effector. Identical controller parameters were applied throughout \textit{the impact phase} and \textit{the traction phase}. Figure~\ref{fig_sim2} displays three sequences of snapshots capturing impact interference: after the non-contact phase (snapshot 1), the impact occurs (snapshot 2), resulting in the tray carrying water being knocked out (snapshots 3-4).

The velocity output curves for the controllers are depicted in Figure~\ref{fig:E_60N_3}.
Under L-AC control, a moderate 40 N impact force resulted in a significant velocity jump of 0.25 m/s (Figure~\ref{fig:E_60N_3}f), leading to water spillage. In contrast, the N-AC exhibited attenuation capabilities for the 40 N impact, with the velocity jump limited to 0.12 m/s (Figure~\ref{fig:E_60N_3}g). However, the attenuation of N-AC becomes saturated as the magnitude of the impact force increases, rendering it unable to resist higher intensity impacts. Under N-AC control, when the impact force reached 70 N, the tray experienced a velocity jump of 0.29 m/s (Figure~\ref{fig:E_60N_3}h), resulting in water spillage.
By comparison, under SFC control, the velocity jump did not exceed 0.1 m/s for both 40 N and 70 N impact forces (Figure~\ref{fig:E_60N_3}i and Figure~\ref{fig:E_60N_3}j), which is one-third of the value for N-AC. This contributed to the smooth movement of the water cup, demonstrating SFC's capability to counteract large impact forces. This result was also confirmed by the end-effector's motion trajectory. In the 70 N impact force test, the robot under N-AC control immediately shifted by 0.08 m, while the robot under SFC control only had a 0.03 m offset, as shown in Figure~\ref{fig:E_60N_3}m and Figure~\ref{fig:E_60N_3}o.

\begin{figure*}[th]
\centering
\includegraphics[width=6.4in]{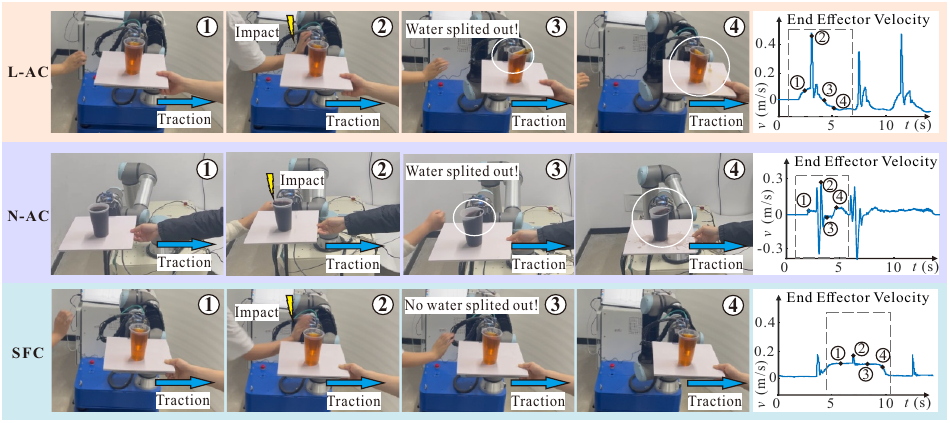}%
\caption{\textit{The simultaneous impact and
traction phase} - experimental snapshots: initialization by human operator, followed by forward traction and impact interference simultaneously. (The duration of the sequence is 5s). The impact and the traction occur simultaneously in the same direction. The blue arrow is the traction force direction.}
\label{snap3}
\end{figure*}
\begin{table*}[th]
  \centering
  \caption{The response of L-AC, N-AC, and SFC to the direction of impact}
    \begin{tabular}{ccccccccc}
    \toprule
    Impact (N) & Traction (N) & Force direction & &\multicolumn{2}{c}{Velocity jump(m/s)} & \multicolumn{2}{c}{Position drift (m)} \\
          &       &       & L-AC  & N-AC  & SFC   & L-AC & N-AC   & SFC \\
    \midrule
    70    & 10    & Same  & 0.5  & 0.3 & \textbf{0.18 } & 0.35 & 0.04 &\textbf{0.01} \\
    10    & 10    & Same  & 0.02  & 0.03 & 0.02  & e-4  & e-3 & e-4 \\
    70    & 10    & Orthogonal  & 0.4  & 0.3 & \textbf{0.1}   & 0.15 & 0.05 & \textbf{0.01} \\
    10    & 10    & Orthogonal  & 0.02  & 0.02 & 0.02  & e-4  &  e-3 & e-4 \\
    \bottomrule
    \end{tabular}%
  \label{tab:Res}%
\end{table*}%
\subsubsection{7.2.3 Simultaneous impact and traction phase (same direction).} Figure~\ref{snap3} exhibits three sequences of snapshots capturing an impact occurring during traction: the robot is being pulled by an operator to transport water (snapshot 1), and suddenly, an impact takes place (snapshot 2), resulting in water being spilled under L-AC and N-AC control; however, under SFC control, the initial motion persists, and the water remains undisturbed (snapshots 3-4).

\begin{figure}[!t]
  \centering
  \includegraphics[width=3.2in]{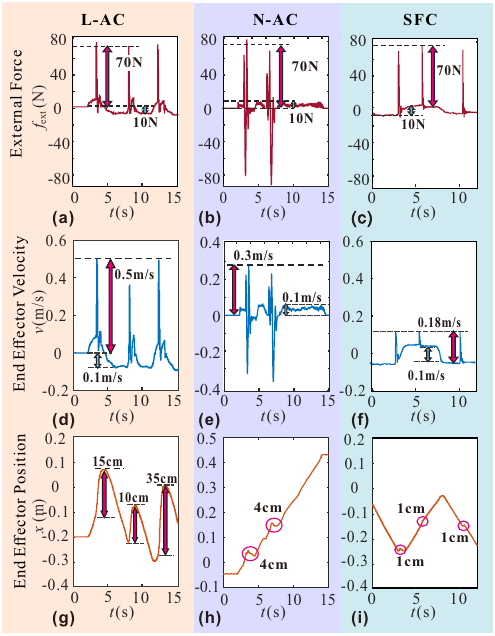}

  \caption{Experimental results of \textit{simultaneous impact and
  traction phase} on fixed manipulator 
  \\
  column 1 (subplots a-d-g):  L-AC - traction and impact  \hspace{1cm} 
  column 2 (subplots b-e-h):  N-AC - traction and impact \hspace{1cm} 
  column 3 (subplots c-f-i):  SFC - traction and impact \hspace{1cm} 
  }
  \label{fig:SFCn_3}
  \end{figure}

During the water delivery process, the operator exerted a force of approximately 10 N to pull the end-effector, while an external force of approximately 70 N impacted the end-effector, as shown in Figure~\ref{fig:SFCn_3}. From Figure~\ref{fig:SFCn_3}d and Figure~\ref{fig:SFCn_3}g, it can be observed that the robot controlled by L-AC experienced a velocity jump of 0.5m/s and a position drift of 35cm upon impact. Consequently, the operator lost control of the robot, leading to water spillage. Although the N-AC system exhibits better resistance to impact forces compared to L-AC, it still resulted in a velocity jump of 0.3m/s and a position drift of 4cm (Figure~\ref{fig:SFCn_3}e and Figure~\ref{fig:SFCn_3}h), also causing water spillage.  In contrast, the robot controlled by SFC experienced a much smaller velocity jump of 0.18m/s and a position drift of only 1cm upon impact (Figure~\ref{fig:SFCn_3}c and Figure~\ref{fig:SFCn_3}i), allowing the operator to smoothly continue dragging the robot. 
The comparative experimental data can also be found in Table \ref{tab:Res}. In addition, it is worth noting two observations. Firstly, the N-AC system exhibited slight oscillations and a reactive force against the operator after being subjected to a forward impact force of 70 N (Figure~\ref{fig:SFCn_3}b, at 0.3 seconds). This phenomenon is expected to be more pronounced in the subsequent experiment. Secondly, when the impact occurred in the same direction as the dragging, the change in position was smaller than when only the impact occurred, which we speculate is due to additional damping introduced by the operator during the dragging.

\subsubsection{7.2.4 Simultaneous impact and traction phase (orthogonal direction).}
\begin{figure*}
  \centering
  \includegraphics[width=6.5in]{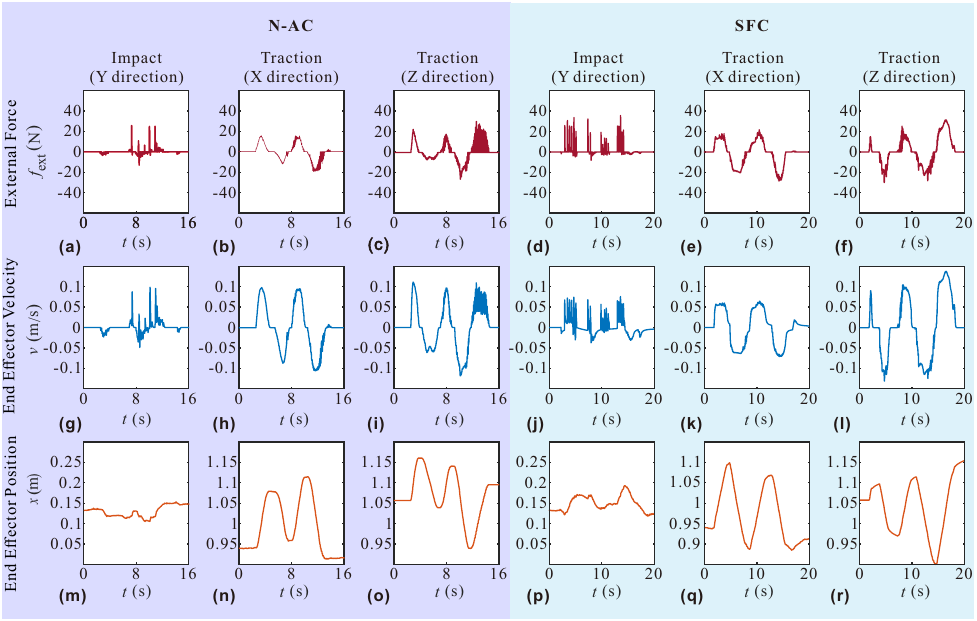}
  \caption{Results of simultaneous impact and traction experiments in the orthogonal direction\\
  \hspace{1.1cm} column 1: (subplots a-g-m) N-AC Moderate impact
  \hspace{1.1cm} column 2: (subplots b-h-n) N-AC Moderate impact\\
  \hspace{1.1cm} column 3: (subplots c-i-o) N-AC Intense impact
  \hspace{1.5cm} column 4: (subplots d-j-p) SFC Moderate impact\\
  \hspace{1.1cm} column 5: (subplots e-k-q) SFC Intense impact
  \hspace{1.5cm} column 6: (subplots f-l-r) SFC Intense impact}
  \label{fig:E_20N_same}
  \end{figure*}

We also tested the case where the impact occurred in the orthogonal direction of traction. Specifically, we applied traction forces of approximately 20 N in the X and Z directions (Figure~\ref{fig:E_20N_same}b, \ref{fig:E_20N_same}c, \ref{fig:E_20N_same}e, \ref{fig:E_20N_same}f), and an impact force of 20 N in the Y direction (Figure~\ref{fig:E_20N_same}a, \ref{fig:E_20N_same}d). Ideally, both N-AC and SFC should resist the impact force of around 20 N. However, under N-AC control, there was an oscillation in velocity in the dragging Z direction with an amplitude of 0.1m/s (Figure~\ref{fig:E_20N_same}c and \ref{fig:E_20N_same}i, at 12 seconds). This is more clearly observed in the video. We speculate that this is due to the damping term in N-AC is directly related to the force, and when disturbances are introduced in other directions, it causes a sudden change in the damping term, which may introduce unexpected nonlinearity and lead to the occurrence of a limit cycle, resulting in self-excited oscillation of the system. 
In contrast, LAC is a linear controller that does not introduce the potential limit cycle associated with nonlinear systems, while SFC's design takes into account the dissipativity of the system, stability of human-robot coupling, and discrete constraints, ensuring stability in force interaction for SFC.

\subsection{7.3 Comparison of results on the mobile manipulator}

Using a lower sampling time can impact the performance of the controller. In order to assess the dynamic response of the SFC  with a slower sampling time ($\Delta T=0.02$ s), we conducted a water transportation experiment involving human-robot collaboration on a mobile manipulator.

\subsubsection{7.3.1 Frequency domain energy analysis of the impact phase.}
\begin{figure*}
  \centering
  \includegraphics[width=6.2in]{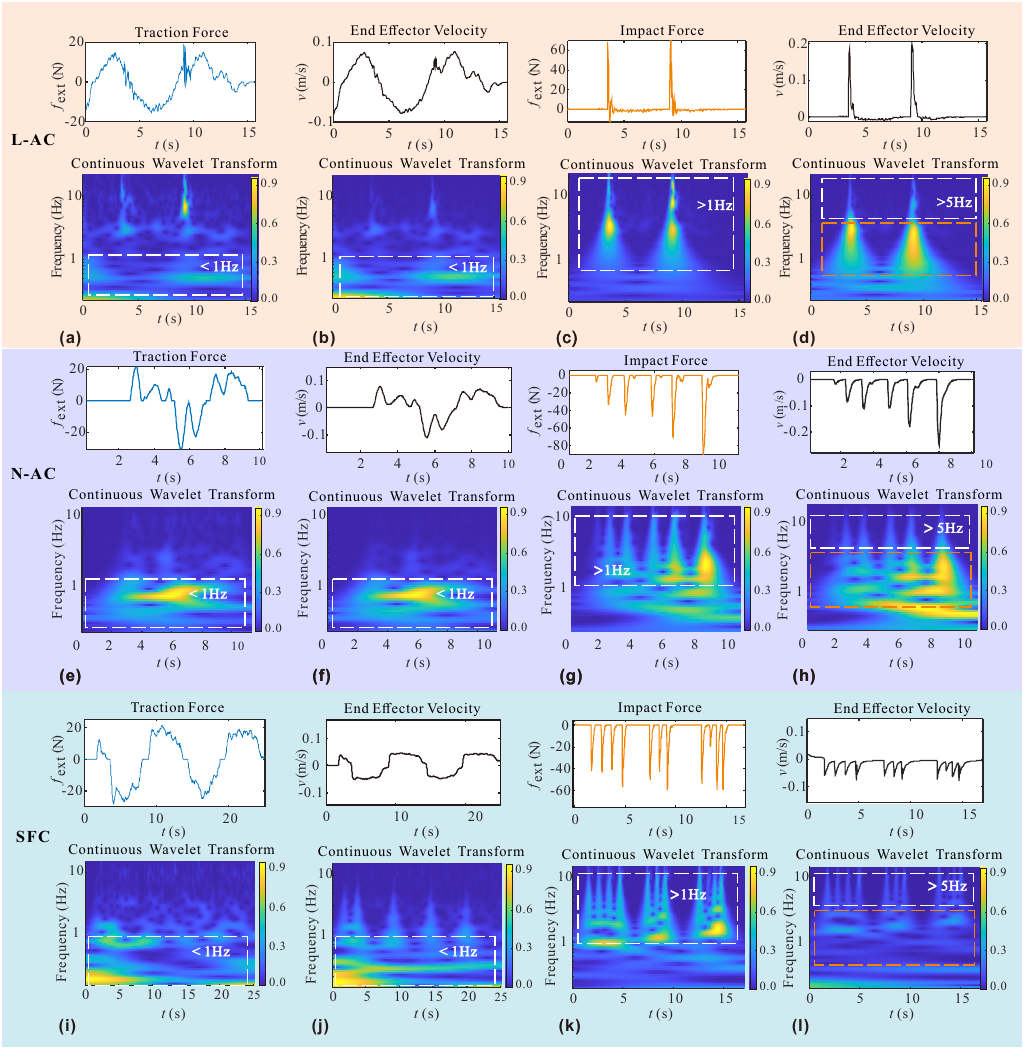}
  \caption{Results of the \textit{impact phase} on the mobile manipulator. The upper half of each subplot illustrates the signals in the temporal domain, while the lower half portrays the associated outcomes of the continuous wavelet transform.\\
  \hspace{1.1cm} row 1: (subplots a-b) L-AC-traction \hspace{5cm}(subplots c-d) L-AC-impact \\
  \hspace{1.1cm} row 2: (subplots e-f) N-AC-traction \hspace{5cm} (subplots g-h) N-AC-impact  \\
  \hspace{1.1cm} row 3: (subplots i-j) SFC-traction  \hspace{5.2cm} (subplots k-l) SFC-impact}
  \label{fig:MMP2}
  \end{figure*}

 L-AC, N-AC and SFC were simultaneously compared from both the time and frequency domains, as depicted in Figure~\ref{fig:MMP2}. The upper half of each subplot represents the signals in the time domain, while the lower half depicts the corresponding continuous wavelet transform results. The wavelet transform was analyzed using Morse wavelets, with the horizontal axis representing time and the vertical axis indicating frequency. In this graph, colors denote energy intensity, exhibiting a normalized representation, where blue to yellow progression corresponds to lower to higher signal energy.

The bandwidth for L-AC, N-AC, and SFC is 8 rad/s (equivalent to 1.3 Hz), as shown in Figure \ref{fig:Exp_bode2}. Therefore, during the application of traction forces (Figure \ref{fig:MMP2}a, \ref{fig:MMP2}e, \ref{fig:MMP2}i), L-AC, N-AC, and SFC all demonstrate the robot's compliance and ensure that signal energy is transmitted within 1 Hz (Figure \ref{fig:MMP2}b, \ref{fig:MMP2}f, \ref{fig:MMP2}j).
When an impact force occurs, a significant accumulation of energy is observed in the frequency range greater than 1 Hz, as shown in Figures \ref{fig:MMP2}c, \ref{fig:MMP2}g, and \ref{fig:MMP2}k. L-AC can only suppress impact energy at frequencies greater than 5 Hz (as shown in Figure \ref{fig:MMP2}d), with the remaining energy causing velocity jumps and position drift in the robot. Although reducing the system bandwidth of L-AC can enhance its ability to resist high-frequency forces, excessively lowering the bandwidth often results in significant delay, thus reducing the comfort of human-robot collaboration. The selected bandwidth for this experiment is already low at 1.3 Hz; however, under L-AC control, impact forces still affect the robot. This finding confirms that while L-AC can effectively handle low-amplitude high-frequency noise, it is not as effective at dealing with moderate- to high-intensity impact forces.
The attenuation in the bode plot of N-AC increases with the increase in input amplitude, but this attenuation is limited. As illustrated in Figure \ref{fig:Exp_bode2}c, the magnitude-frequency response curve of N-AC overlaps for impact forces exceeding 50 N. This indicates that N-AC has weak resistance to larger impact forces. 
Figure \ref{fig:MMP2}h further demonstrates this point by showing that although N-AC dissipates impact energy above 5 Hz, it does not effectively dissipate energy within the 5 Hz range and even exhibits some enhancement. We speculate that this might be one of the reasons causing self-excited oscillations in the N-AC system. In contrast, even at a low control frequency, SFC significantly attenuates the transmission of impact energy under the same bandwidth, with nearly complete attenuation of impact energy above 1 Hz (as shown in Figure \ref{fig:MMP2}f).

  \begin{figure}[!t]
    \centering
    \includegraphics[width=3.2in]{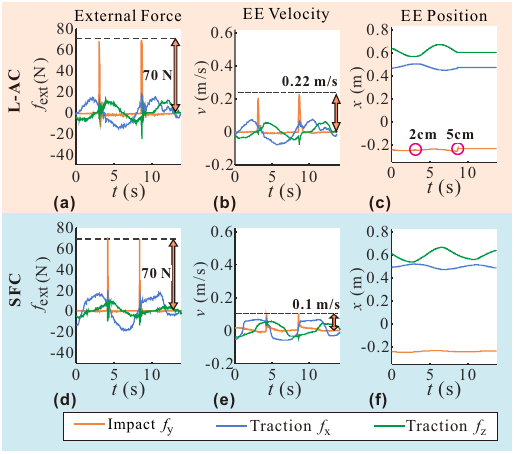}
    \caption{Experimental results of \textit{simultaneous impact and
    traction phase} on mobile manipulator  \\
    \hspace{1.1cm} row 1: (subplots a-b-c) L-AC-traction and impact  \\
    \hspace{1.1cm} row 2: (subplots d-e-f) SFC-traction and impact }
    \label{fig:MMP}
  \end{figure}

\subsubsection{7.3.2 Simultaneous impact and traction phase}
In this experiment, traction forces were applied in the X and Z directions, while impact forces were applied in the Y direction. Moreover, considering the results from Section 7.2.4, where simultaneous dragging and impact caused self-excited oscillations in N-AC, which is dangerous for a 370 kg mobile platform, we only compared the cases of simultaneous dragging and impact under L-AC and SFC on the mobile manipulator.
The experimental results, shown in Figure \ref{fig:MMP}, demonstrate that both SFC and L-AC achieve collaboration with the traction force. When an impact occurs, the velocity jump in SFC is less than 0.1 m/s, with minimal position drift. In contrast, under L-AC, the velocity jump is twice that of SFC, with 5 cm of position drift. Therefore, even at a lower control frequency, SFC exhibits superior resistance to impacts compared to L-AC.

\subsection{7.4 Discussion}

The real-world experiments compared the performance of L-AC, N-AC and SFC in the context of human-robot collaboration. The robot collaborated with a human to deliver water to an arbitrary destination through force perception, a process that is disturbed by uncertain impact. These experiments were conducted on two robotic platforms: a fixed manipulator and a mobile manipulator. Based on the above physical experiments, the following conclusions can be drawn.

1) On the fixed manipulator, SFC exhibited better impact resistance capability compared to both L-AC and N-AC. Experiments were conducted during the traction phase, impact phase, and simultaneous traction and impact phases under different intensities, multiple dimensions, and various orientations. 
The results showed that both SFC and N-AC demonstrated a compliant effect similar to that of L-AC, with SFC even conserving more effort during the traction phase. When subjected to moderate-intensity impacts, both SFC and NAC exhibit the ability to suppress the impact. However, as the magnitude of the impact force increases, the attenuation of NAC tends to saturate, while SFC continues to attenuate. This allows SFC to better suppress stronger impact disturbances compared to NAC. In the presence of significant impacts, the velocity jump and position drift of the robot under SFC are significantly smaller compared to those under L-AC and N-AC. Furthermore, in the event of impact disturbances during the traction phase, a robot controlled by SFC effectively suppresses the impact effects and maintains good collaboration with the operator. This can be attributed to the dissipative properties considered in the design of SFC, as well as the discrete constraints and stability of human-robot interaction coupling. However, when impacts and traction occur simultaneously, the N-AC occasionally exhibits self-excited oscillations. We speculate that the self-excited oscillations may be attributed to the nonlinear damping characteristics of N-AC, where the nonlinear damping is explicitly dependent on force, and sudden changes in force lead to sudden changes in damping, potentially introducing limit cycles and self-excited oscillations. Therefore, a robot controlled by SFC can better withstand impact disturbances and maintain good collaboration with the operator.

2) On the mobile manipulator, SFC was verified to achieve impact resistance even at low control frequencies. The control frequency of the mobile manipulator was only 50~Hz, one-tenth of that of the fixed manipulator. We adapted the parameters based on the discrete control constraints proposed in Section~6, and the experimental results showed that the mobile manipulator could still complete the human-robot collaborative water delivery task under SFC without spilling the water. Additionally, the wavelet transform was used to compare the energy distribution in the time and frequency domains of the robot response under L-AC, N-AC and SFC. Even though the system bandwidth of the L-AC and N-AC were designed to be very narrow, they still could not completely suppress the impact energy. Although L-AC could suppress noise at high frequency, it was challenging to attenuate high-intensity impact at medium frequency. In contrast, SFC significantly attenuated the impact energy, ensuring that the robot could still comply with the human operator to complete the collaborative task when impacts occurred.

\section{8. Qualitative case study}
\begin{figure*}[!t]
\centering
\includegraphics[width=6in]{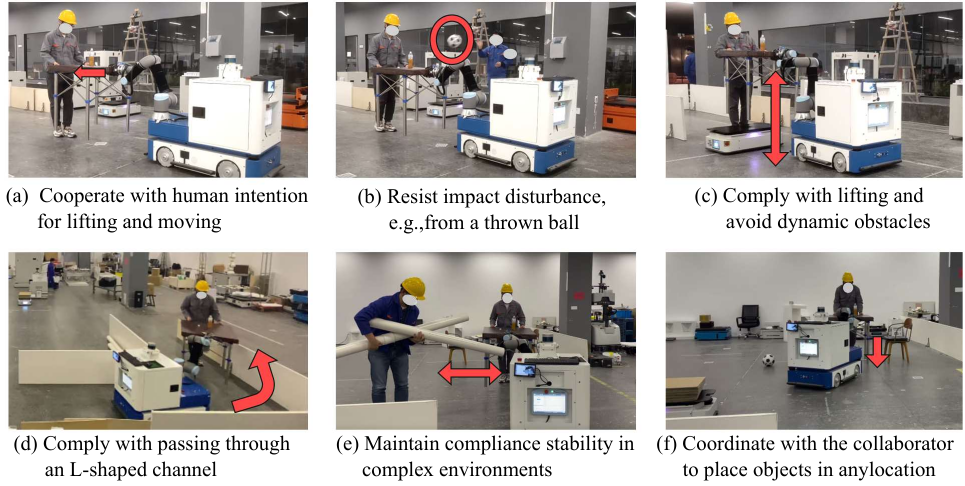}%
\caption{Efficient collaboration and impact resistance of mobile manipulator robot with SFC in dynamic factory environment}
\label{fig:case}
\end{figure*}

This case study demonstrates the collaboration process of a mobile manipulator robot controlled by SFC. The robot works with a human worker to lift a table with an open water bottle on it. With force sensing, the robot complies with the worker's intention and performs a series of transportation and obstacle avoidance tasks. During this process, there may be uncertain external impact disturbances (e.g. from a thrown ball) that the robot must resist to ensure that the water on the table does not spill. Finally, the robot collaborates with the human worker to place the table at a designated location while ensuring comfortable, smooth, and safe collaboration. The complete procedure is demonstrated in Extension 2. Figure \ref{fig:case} offers visual depictions of these performance aspects.


The study focuses on the qualitative performance aspects of a mobile manipulator robot controlled by SFC. The robot's ability to cooperatively handle tasks requiring physical exertion with a human operator (Figure \ref{fig:case}a), its robustness in resisting impact disturbances from thrown objects (Figure \ref{fig:case}b), and its capability to adapt to changes in lifting requirements while circumventing obstacles (Figure \ref{fig:case}c) are all demonstrated. Furthermore, the robot exhibits robustness in its ability to sense and respond to contact forces, when navigating through challenging environments such as L-shaped channels (Figure \ref{fig:case}d), and maintains stability despite environmental complexity (Figure \ref{fig:case}e), highlighting its adaptability and control system reliability. Finally, the robot collaborates effectively with a human partner, showing potential for robots to enhance human productivity through flexible and reliable placement of objects (Figure \ref{fig:case}f). 


Overall, this case study illustrates the capability of a mobile manipulator robot to comply with worker's intentions for transporting the table and resist external perturbations by utilizing SFC. The study highlights the potential for mobile robots to enhance human capabilities in tasks that require physical interaction and cooperation. Furthermore, it provides valuable insights into the design and implementation of such systems for practical applications in diverse settings.

\section{9. Related work}
In the field of pHRI, the control of robots during interaction tasks is an active area of research. There are several key areas of pHRI control that are relevant to the development of a controller that enables robots to resist impact forces while maintaining compliance with traction forces. These areas are admittance control parameters tuning, nonlinear damping control, and the technology of distinguishing between impact forces and intentional contacts.
In this section, we will discuss these areas of pHRI control and summarize the related research progress.

\subsection{9.1 Admittance control parameters tuning}
Admittance control, as favored by \cite{chen2022human}, is a preferred approach in physical human-robot interaction (pHRI) due to its ability to allow robots to move based on contact forces and comply with the operator's intentions. 
Compared with impedance control, admittance control is more beneficial for non-backdrivable or challenging-to-identify dynamic parameter robots to achieve flexible interaction, as noted by both~\cite{chen2020compliance,arduengo2021human}.  Tuning admittance control parameters is crucial for optimizing the robot's response during pHRI, and this compliant interaction response mainly reflects three aspects: stability, passivity, and transparency.

1) In the context of stability, researchers have conducted valuable studies from the perspectives of frequency domain detection, feedforward control, velocity bandwidth, virtual damping augmentation, and energy allocation. They have provided beneficial design principles and adaptive adjustment strategies (\cite{dimeas2016online,keemink2018admittance,ferraguti2019variable}). However, these analyses primarily focus on tightly coupled human-machine interactions, with limited research on contact transitions and the impact of impulsive forces. 2) Passivity is an effective approach for analyzing the stability of complex systems. Preserving the power envelope, unaffected by the passivity condition, prevents mismatches in admittance parameters caused by unpredictable human interaction (\cite{ferraguti2015energy}). Motivated by the concept of passivity, our paper investigates the dissipation of the proposed controller under impact disturbances to ensure the safety of human-robot interaction. 3) To achieve transparency in pHRI, the robot should minimize its resistance to human motion (\cite{laghi2020unifying}). Researchers have explored methods such as fractional-order control and integrating admittance control with arm redundancy resolution to enhance transparency (\cite{sirintuna2020variable,aydin2021towards,kim2012admittance} ). However, transparency is not always the optimal solution in pHRI. In situations where impact disturbances occur, high transparency can lead to sudden changes in robot motion speed, posing a safety threat to collaborators. In such cases, lower transparency may be preferable to avoid unnecessary secondary injuries. 

When designing pHRI systems, researchers have mainly focused on the trade-off between passivity, stability, and transparency in admittance control's dynamic properties. However, distinguishing between impact forces and intentional contacts has not been addressed from a dynamical perspective. In fact, an admittance controller with linear inertial-damping-stiffness formulation does not differentiate forces of different amplitudes and frequencies. Therefore, this paper considers nonlinear control to achieve the desired dynamic properties while meeting the requirements for passivity and stability.

\subsection{9.2 Nonlinear damping control}
Nonlinear damping control has been proposed as an effective method for regulating the force interaction between a robot and a human. 
Incorporating nonlinear damping in control system design can improve error dynamics convergence, counteract estimation errors, achieve robust control, reduce overshoot, and ensure stability, passivity, convergence time, and accuracy for second-order systems (\cite{6202693,7954629,lai2014improving,ruderman2021optimal}). In recent years, variable damping controllers have gained significant attention in research. The focus of these studies includes addressing changes in system dynamics caused by unknown disturbances, balancing stability and agility, minimizing energy consumption (\cite{6809166,zahedi2021variable,zahedi2022user}).

Compared to linear cases, nonlinear damping control offers better robustness in uncertain systems. However, designing such controllers is complex and typically involves system modeling, simulation, and parameter tuning tailored to specific applications, which poses technical difficulties and high workload. While some biologically inspired parameter regulation methods, such as imitating human muscle stiffness (~\cite{al2018active}) or implementing an angle-related damping curve for impact-bearing joints (~\cite{hamid2021state}), have been proposed, they often aim to follow contact forces instead of specific dynamic properties. Moreover, since nonlinear damping control methods may contain multiple adjustable parameters, the tuning and experimental verification process can be time-consuming. Thus, automatic parameter tuning methods for nonlinear controllers are necessary to alleviate workload associated with manual tuning while ensuring optimal performance.


Researchers have extensively studied nonlinear damping and its stability analysis using various approaches, such as compliant contact models, quasi-linear modeling methods, energy-based approaches, and Lyapunov functionals (\cite{61008,798060,marx2019stability,colonnese2016stability,elliott2015nonlinear}). These studies have provided valuable insights into the performance, stability, and practical applications of nonlinear damping models. However, these models do not meet our specific requirements for dynamic compliance with traction and resistance to impacts. To address these limitations observed in linear models, this paper analyzes the performance of a proposed nonlinear controller, drawing from the aforementioned studies to gain a better understanding of the system's behavior.

\subsection{9.3 Distinguish between impact and compliance}

Distinguishing between impact and compliance is critical in pHRI tasks as it enables robots to differentiate between unintended collisions and intentional contact. One approach is to use force sensing technologies to detect the level of force during interaction tasks. \cite{li2018stable} proposed a method that combines variable admittance control and adaptive control, utilizing two independent force sensors. The objective is to maintain human-like compliance in direct control of the robot while ensuring smooth transitions and stable motion when the robot interacts with the environment. Other approaches address impact and compliance issues through mechanical design, including adjusting the stiffness of parallel compliant elements, utilizing passive compliant links and joints, or designing variable stiffness mechanisms (\cite{niehues2015compliance,she2020comparative,ayoubi2020safe}). However, it is crucial to note that compliant mechanisms may suffer from hysteresis, fatigue, and creep effects over time, which can lead to performance degradation.


Energy observation-based methods are also common in addressing impact disturbance, where real-time model-based collision detection, isolation, and recognition can be achieved through energy observers and momentum observers \cite{haddadin2017robot}. \cite{lachner2021energy} allocated a safe energy budget for robots and modifies parameters to restrict the exchange of kinetic and potential energy during collisions, allowing automatic tuning of controller parameters and reducing impact (\cite{raiola2018development}, \cite{munoz2019time}, \cite{navarro2016iso10218}). In unstructured environments, some studies focused on detecting collisions and ensuring robots react quickly to prevent potential hazards (\cite{de2012integrated}, \cite{khan2014compliance}). Nonetheless, techniques like math model matching or signal threshold filtering may introduce time delays and recovery errors, ultimately affecting the real-time responsiveness and accuracy of pHRI (\cite{haddadin2017robot}, \cite{lin2021adaptive}). Our SFC controller does not require online identification of external forces. Instead, it achieves different responses to different forces through the inherent nonlinearity of the controller, which contributes to better real-time performance.

\subsection{9.4 Inverse kinematics}
Inverse kinematics is a crucial aspect in pHRI. However, the computation of the inverse Jacobian matrix can present safety issues in terms of singularities and redundancy management. Impedance control is an effective approach for addressing singularities as it maps end-effector forces to joint forces using the transpose of the Jacobian matrix, thereby avoiding the issues associated with inverting the Jacobian (\cite{albu2002cartesian}). Nonetheless, impedance control also has some limitations. Uncertainties in the robot's dynamic parameters and noise in acceleration measurements can have a detrimental impact on the accuracy of impedance control (\cite{lynch2017modern}). Moreover, impedance control encounters difficulties in generating stiff virtual dynamic models (\cite{keemink2018admittance}) and is not feasible for non-backdrivable and high-inertia robots (\cite{villani2016force}). 

In addition to impedance control, there exist multiple alternative methods for addressing singularities and managing redundancy in inverse kinematics. For example, the weighted pseudo-inverse Jacobian method (\cite{mussa1991integrable}) or the damped least squares method (\cite{zhan2021adaptive}) can be utilized to effectively deal with singularity problems in inverse kinematics. Another strategy involves decoupling the mobile platform from the robot manipulation system to address redundancy challenges (\cite{chung1998interaction}). Moreover, employing null-space projection techniques enables dynamic control of redundant manipulators, enabling them to effectively execute multiple priority tasks concurrently (\cite{ott2015prioritized}). In our case, we employed a damped least squares approach to calculate the pseudo-inverse of the Jacobian. When approaching singularity, we only apply small damping along the singular directions to reduce the impact of the singularity. It is important to note that inverse kinematics is not the main focus of this paper, hence the limited discussion on this topic.

To sum up, pHRI is an area of ongoing research, and the studies reviewed in this section have demonstrated promising outcomes in enhancing the efficiency, safety, and performance of human-robot interaction through creative control methods. However, to achieve more natural and secure interactions, it is necessary for robots to withstand external impact disturbances while complying with human traction. Currently, this dynamic ability has not been attained through the aforementioned research. With inspiration from shear-thickening fluids, there is potential to design a controller that can achieve the desired dynamic characteristics of pHRI.


\section{10. Conclusion}
Inspired by the phase change properties of shear-thickening fluids, this paper proposes an SFC that can resist impact disturbances while complying with the traction during the human-robot collaboration. 
SFC is a nonlinear controller where system response depends on both frequency and amplitude of the input signal. 
In this paper, we analyze in detail the stability, passivity, and phase trajectory of SFC to ensure its feasibility. Specifically, we analytically derive SFC's frequency domain response and its dynamic properties, including the system bandwidth, time constant, and system gain variation. Such analysis not only provides theoretical support for explaining SFC's ability to achieve traction compliance and impact resistance but also provides a framework for parameter settings. Moreover, given the existing robotic systems' discrete digital control, the instability problem induced by the low control frequencies must be considered. Therefore, we further analyze the acceleration convergence of SFC in iteration and provide parameter constraints under discrete control to guarantee stability in real human-robot collaboration.
We compared the frequency and time domain characteristics of L-AC, N-AC, and SFC through simulations, and validated the dynamic properties and parameter constraints of SFC in discrete control. In real-world experiments, we compared the performance of L-AC, N-AC, and SFC in both fixed and mobile manipulators. SFC demonstrated superior impact resistance and maintained stable collaboration, making humans more comfortable in collaboration.

The proposed SFC ensures safe human-robot collaboration by resisting external impact interference while being compliant with human traction. The SFC has been integrated into industrial collaborative handling and has potential applications beyond human-robot collaboration. We will investigate the use of SFC in exoskeletons for protecting the wearer from impact forces while transparently complying with their movement intentions. Furthermore, as shear-thickening fluid is only one form of non-Newtonian fluid, we will explore other diverse hydrodynamic properties in future research for application in exoskeletons, teleoperated robots, and rehabilitation robots.

\begin{funding}
  This work was supported in part by the National Nature Science Foundation of China under Grant 62173293 (R.X.), 62373322 (Y.W.), and 62303407 (H.L.); in part by the “Ling-Yan” Research and Development Project of Zhejiang Province 2023C01176 (Y.W.); in part by the Zhejiang Provincial Natural Science Foundation of China LD22E050007 (H.L.); in part by the Innovation and Development Special Fund of the Hangzhou Chengxi Sci-tech Innovation Corridor (R.X.); in part by the Tencent AI Lab Rhino-Bird Focused Research Program JR201987.
\end{funding}

\bibliographystyle{SageH}
\bibliography{reference/reference}


\newpage
\appendix

\section{Appendix A: Index to multimedia extensions}
The relevant extension videos associated with this paper are shown in Table \ref{tab:multimedia}.

\begin{table}[!ht]
\small\sf\centering
\caption{Multimedia Extensions\label{tab:multimedia}}
\begin{tabular}{lll}
\toprule
Extension & Type & Description\\
\midrule
\texttt{1}& Video & The compliance and impact resistance of  \\ 
         &  &  L-AC, N-AC, and SFC have been verified   \\
         &  & through real world experiments. \\
\texttt{2}& Video & The performance of the proposed SFC \\
 &  & was evaluated throughout the industrial \\  
 &  & collaborative handling process. \\
\bottomrule
\end{tabular}
\end{table}

\section{Appendix B: From constitutive equation to virtual dynamics}
\begin{figure}[!t] 
    \centering
    \includegraphics[width=8cm]{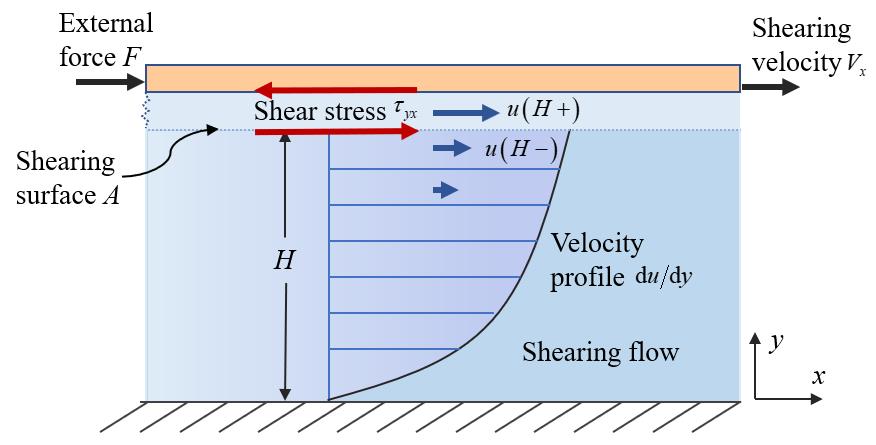}
    \caption{Fluid mechanics: a simple shearing. The x-direction is the direction of fluid flow and the y-direction is the depth of fluid.}
    \label{fig:shearing_flow}
\end{figure}

Fluid dynamics analysis usually assumes that the shear-thickening fluid flows along the $x$-direction with a velocity of $u$ and only has a velocity gradient along the $y$-direction~\cite{painter2019fundamentals}. As shown in Figure~\ref{fig:shearing_flow},
consider a thin layer of flow confined between two wide parallel plates separated by a small distance $H$. The bottom plate keeps stationary, while the top plate is free to move. 
Suppose a constant external force FF is applied to the top plate. Under a steady state the top plate moves at a shear velocity ${V}_x$, which gives rise to an equal but opposite internal frictional force in the fluid on account of the fluid viscosity. 
The local velocity $u\left( y \right)$ increases in the positive direction along the y-axis, the velocity ($u\left( H+ \right)$), where it is slightly higher than plane $A$, is higher than that ($u\left( H- \right)$) under plane $A$. Thus, a ``pull" action is formed, trying to pull the fluid micromass to the positive direction of the $x$-axis.
Therefore, shear stress $\tau_{yx}$ is applied to a shear surface $A$ perpendicular to the $y$-axis.
From a fluid mechanics viewpoint, the shear stress $\tau_{yx}(y)$ in the flow is positively correlated with the local velocity gradient~$\text{d}u/\text{d}y$.

In the field of fluid mechanics, fluid dynamic properties are often expressed in the form of constitutive equation~\cite{zatloukal2010simple,ferras2012analytical,perlacova2015tensorial,zatloukal2020frame}. We study the shear-thickening fluid in terms of its constitutive equation.

\begin{equation}   \label{eq:SimpleConstitutiveEquation}
{\tau _{yx}} = \mu {\left| {\frac{{{\rm{d}}u}}{{{\rm{d}}y}}} \right|^{n{\rm{ - }}1}}\frac{{{\rm{d}}u}}{{{\rm{d}}y}}
\end{equation} 
where ${\tau _{yx}}$ represents the shear stress component acting on the plane perpendicular to the $y$-direction, and its direction is $x$-direction. $\mu \ge 0$ is the apparent viscosity, a material characteristic describing the internal flow resistance. $n>1$ is a power-law term for shear thickening fluids, introducing the nonlinear characteristics.

We build an appropriate dynamic equivalent model from shear-thickening fluid and incorporate it into the control system. 
In this way, the controlled object behaves like a shear-thickening fluid with dynamic properties of \textit{ compliance with traction and resistance to impacts}.
The dynamics equation of the controller is obtained by analogy from the constitutive equation of the fluid (\ref{eq:SimpleConstitutiveEquation}). Replacing $\frac{{{\text{d}}u}}{{{\text{d}}y}}$ with ${\dot x}$ and ${\tau _{yx}}$ with ${f_\text{ext}}$, we get the dynamic equation for the controller viscosity term in the form of a shear-thickening fluid:
 \begin{equation} \label{eq:ControlKernel}
     {f_\text{ext}} =D(\dot x)= \mu {\left| {\dot x} \right|^{n{\rm{ - }}1}}\dot x
 \end{equation}
 where $f_\text{ext}$ represents the external force on the object, and ${\dot x}$ means the velocity of the controlled object.

\section{Appendix C: Properties of the SFC} 

\subsection{Property 1: Non-linear spatial velocity ratio.}
Phase trajectory refers to the path that the robot's position and velocity follow over time. Analyzing the phase trajectory can provide visual representations of the state space of a robot controller, enabling us to observe the convergence of trajectories in the phase plane and obtain an intuitive understanding of the system's dynamic response characteristics. The phase plane where the phase trajectory is located is a two-dimensional plane, where the horizontal and vertical axes represent two variables in the system state. Typically, the horizontal axis $x_1$ represents the position state, and the vertical axis $x_2$ represents the velocity state. 

The state equation of the system of the form (\ref{eq:with_mass}) can be expressed as:
(\ref{eq:state_equation}), 
\begin{equation}\label{eq:state_equation}
\left\{ {\begin{array}{*{20}{c}}
  {{{\dot x}_1} = {x_2}} \\ 
  {{{\dot x}_2} = \frac{{ - \mu {{\left| {{x_2}} \right|}^{n - 1}}{x_2}}}{m}} 
\end{array}} \right.
\end{equation}

 The spatial velocity ratio provides information about the relative magnitudes of the state change in the two directions at that particular state on the phase trajectory. If $x_1$ represents the position state and $x_2$ represents the velocity state, a larger spatial velocity ratio indicates that the rate of change of the velocity state $\frac{{{\text{d}}{x_2}}}{{{\text{d}}t}}$ is relatively faster than the rate of change of the position state $\frac{{{\text{d}}{x_1}}}{{{\text{d}}t}}$ at a specific point on the phase trajectory. This means that under the same position changes, the magnitude of velocity changes is greater.

The spatial velocity ratio of its phase trajectory in the phase plane is determined by
\begin{equation} \label{eq: dx2/dx1}
{R_\text{sv}} = \frac{{{\text{d}}{x_2}}}{{{\text{d}}t}}\frac{{{\text{d}}t}}{{{\text{d}}{x_1}}} = \frac{{{{\dot x}_2}}}{{{{\dot x}_1}}} =  - \frac{\mu }{m}{\left| {{x_2}} \right|^{n - 1}}
\end{equation}

According to equation (\ref{eq: dx2/dx1}), we can determine the relative magnitudes of the spatial velocity ratios for linear SFC system with $n=1$ and nonlinear SFC system with $n>1$, when the velocity $x_2$ is greater than 1.

\begin{equation}\label{key}
    {\left. {{R_\text{sv}}} \right|_{{x_2} \geqslant 1,n > 1}} \geqslant {\left. {{R_\text{sv}}} \right|_{{x_2} \geqslant 1,n = 1}}
\end{equation}

Therefore, when the velocity is high (${x_2} \geq 1$), the nonlinear system ($n>1$) has a larger spatial spatial velocity than the linear system($n=1$). Similarly, the converse can be proven.

\subsection{Property 2: Stability }
The stability of a robot controller refers to the controller's ability to maintain the robot's stable state during task execution, which is crucial for avoiding unexpected movements or loss of control. In the case of the SFC controller proposed in this paper, it is necessary to ensure stable velocity tracking of the robot, similar to a cruise control system, which enables the robot to perform tasks such as human-robot interaction involving dragging.

Consider a system of the form (\ref{eq:with_mass}), assuming $M>0, \mu>0, n>0$ and ${f_{\text{ext}}}$ is a consistent input. Then, there is a unique equilibrium point at  $\dot x = {\dot x^ * }$, which is globally asymptotically stable.

\begin{proof}
For any consistent input ${f_{\text{ext}}}$, there exists a unique solution ${{\dot x}^ * } = {\left| { \frac{f_\text{ext}}{\mu}  } \right|^{{\frac{1}{n}}}}{\rm{sign}}\left( {{f_\text{ext}}} \right)$ such that the derivative of $\dot x\left( t \right)$ with respect to time, $\ddot x ={ \frac{f_\text{ext}}{m}} - {\frac{\mu}{m}}{\left| {{{\dot x}^ * }} \right|^{n - 1}}{{\dot x}^ * }$, is equal to zero.  
To simplify the analysis of the system, let us define a new variable as $s=\dot x - {\dot x^ * }$ that shift the equilibrium point to the origin. Then the original differential equation can be expressed in terms of $s$
\begin{equation}  \label{eq:s}
    m\dot s + \mu {\left| s \right|^{n - 1}}s = 0
\end{equation} 

Take the Lyapunov function
\begin{equation}
	V\left( {s} \right) = {s^2}
\end{equation}

It is evident that $V\left( {s} \right)$ is positive definite ($s \in \mathbb{R}$) and radially unbounded. According to (\ref{eq:s}), the derivative of the Lyapunov function along the system trajectory  can then be derived as
\begin{equation} \label{eq:V_d}
\dot V\left( s \right) = 2s\dot s =  - {\frac{2\mu {{\left| s \right|}^{n - 1}}{s^2}}{m}}
\end{equation}

From the given conditions $M>0$, $\mu>0$ and $n>0$, it is clear that $\dot V\left( {s} \right)$ is negative definite. Therefore, according to the Lyapunov stability theory, the system is globally asymptotically stable.
\end{proof}

SFC is a velocity controller that ensures the velocity state of the system converges to the equilibrium point over time. As for the position state, it will ultimately fall onto the singular line of $\dot x = {{\dot x}^ * }$, as confirmed earlier during the phase trajectory analysis.

\subsection{Property 3: Passivity }
To ensure the reliability of a robot system during task execution, its controller must consider the interaction between the robot and the environment, and manage energy conversion and dissipation. We previously analyzed the stability of the system under constant external forces. However, to evaluate the stability of the system under other external disturbances, we use dissipativity analysis. A robot controller with passivity characteristics can prevent unrestricted energy accumulation during the control process, thus ensuring the system's reliability.

Consider a dynamical system described by the input-output equation (\ref{eq:with_mass}), where $f_\text{ext}$ is the input, and $\dot x$ is the output. Assume that the system parameters satisfy $M>0$, $\mu>0$, $n>0$. Then the system is passive and dissipative.

\begin{proof}
Let $V(\dot x) = {\frac{1}{2}}M{{\dot x}^2}$ be a Lyapunov-like function. Then the system can be expressed in a power form as follows:
\begin{equation} \label{eq:g}
\frac{d}{{dt}}\left( {V\left( {\dot x} \right)} \right) = m\dot x\ddot x = \dot x{f_\text{ext}} - g\left( {\dot x} \right)
\end{equation}
where $V(\dot x) = \frac{1}{2}M{\dot x}^2$ is the total energy stored in the system, and $g(\dot x) = \mu {\left| {\dot x} \right|^{n-1}}{{\dot x}^2}$ is the dissipated power. 

As $g(\dot x) \geq 0$ and $\frac{1}{2}M{\dot x}^2$ has a lower bound, we can conclude that the change in stored energy is less than or equal to the energy provided, implying that the system is dissipative. Moreover, as the dissipated power is always non-negative, the system is passive. Hence, the input-output mapping between ${f_\text{ext}}$ and $\dot x$ is passive.

Moreover, a system is said to be dissipative if the change in stored energy is less than or equal to the energy provided~\cite{xia2016passivity}. Since $\int_0^\infty  {\dot x{f_\text{ext}}} {\text{d}}t \ne 0$, and $\int_0^\infty  {g(\dot x)} {\text{d}}t > 0$, thus the passive system in (\ref{eq:g}) is dissipative, and $g(\dot x)$ is the power consumed. 
\end{proof}

\section{Appendix D: Frequency domain analysis of~SFC}
Assuming that the output of the SFC system described in Equation (\ref{eq:with_mass}) is a sinusoidal signal given by $\dot x = B\sin(\omega t)$, the damping term of the controller can be expressed as 
\begin{equation}
    \mu {\left| {\dot x} \right|^{n - 1}}\dot x = \mu {\left| {B\sin (\omega t)} \right|^n}{\text{sign}}\left( {B\sin (\omega t)} \right)
\end{equation}

Because $\dot x = B\sin ( {\omega t} )$ is an odd periodic function with period $T = \frac{{2\pi }}{\omega }$, and $\mu {\left| {\dot x} \right|^{n - 1}}\dot x$ is also an odd function. By using Fourier series expansion, we can express $\mu {\left| {\dot x} \right|^{n - 1}}\dot x$ as:
\begin{equation}\label{eq:Fourier}
\begin{gathered}
  \mu {\left| {\dot x} \right|^{n - 1}}\dot x = \sum\limits_{m = 1}^\infty  {{b_m}\sin (m\omega t)}  \hfill \\
  {b_m} = \frac{2}{T}\int_0^T \mu  {\left| {B\sin (\omega t)} \right|^n}{\text{sign}}\left( {B\sin (\omega t)} \right)\sin (m\omega t){\text{d}}t \hfill \\ 
\end{gathered} 
\end{equation}

To simplify the Fourier series expansion of the damping term in the SFC system, it is observed that the linear part of the system acts as a low-pass filter, thus the higher harmonics with $m>1$ in (\ref{eq:Fourier}) can be neglected, leaving only the fundamental signal ${b_1}\sin (\omega t)$. By expressing the resulting ${b_1}$ integral in terms of the Gamma function, the expression can be further simplified as

\begin{equation}
{b_1} = \mu {B^n}\Psi \left( n \right)
\end{equation}
where $\Psi \left( n \right) = \frac{{2\sqrt \pi  \Gamma \left( {1 + \frac{n}{2}} \right)}}{{\Gamma \left( {\frac{{3 + n}}{2}} \right)}}$, which is a function related only to the power-law $n$. And $\Gamma \left( x \right) = \int_0^{ + \infty } {{t^{x - 1}}} {e^{ - t}}dt\left( {x > 0} \right)$ denotes the Gamma function, which is also known as Euler's second integral.

Thus, an approximation of the damping term $\mu {\left| {\dot x} \right|^{n - 1}}\dot x$ can be obtained as:
\begin{equation} \label{eq:b1}
\mu {\left| {\dot x} \right|^{n - 1}}\dot x \approx \mu {B^n}\Psi \left( n \right)\sin \left( {\omega t} \right)
\end{equation}

The corresponding system input $f_{{\rm{ext}}}$ can be expressed as
\begin{equation}
\begin{gathered}
  {f_{{\text{ext}}}} = m\ddot x + \mu {\left| {\dot x} \right|^{n - 1}}\dot x \hfill \\
   \approx MB\omega \cos \left( {\omega t} \right) + \mu {B^n}\Psi \left( n \right)\sin \left( {\omega t} \right) \hfill \\
   = A\sin \left( {\omega t + \phi } \right) \hfill \\ 
\end{gathered}
\end{equation}
where the amplitude of the input force is
\begin{equation} \label{eq:A(B,w)}
A(B,\omega ) = \sqrt {{{\left( {mB\omega } \right)}^2}{\text{  +  }}{{\left( {\mu {B^n}\Psi \left( n \right)} \right)}^2}} 
\end{equation}
and the input phase is
\begin{equation}
\phi (B,\omega ) = \arctan \left( {\frac{{mB\omega }}{{\mu {B^n}\Psi \left( n \right)}}} \right)
\end{equation}

Finally, we can obtain the describing function of the SFC system, which represents the ratio of the standard sinusoidal output to the fundamental component of the input.
\begin{equation}\label{eq:aa}
\begin{gathered}
  N(B,\omega ) = \frac{{B{e^{j\omega t}}}}{{A{e^{j(\omega t + \phi )}}}} = \frac{B}{A}{e^{ - j\phi }} \hfill \\
   \approx \frac{B}{{\sqrt {{{(MB\omega )}^2}{\text{   +   }}{{\left( {\mu {B^n}\Psi \left( n \right)} \right)}^2}} }}{e^{ - j\arctan \left( {\frac{{mB\omega }}{{\mu {B^n}\Psi \left( n \right)}}} \right)}} \hfill \\ 
\end{gathered}    
\end{equation}

Moreover, the amplitude frequency response of the SFC system can be expressed as
\begin{equation} \label{eq:AF}
\left| {N\left( {B,\omega ,n} \right)} \right| \approx \frac{1}{{\sqrt {{{\left( {m\omega } \right)}^2}{\text{  +  }}{{\left( {\mu {B^{n - 1}}\Psi \left( n \right)} \right)}^2}} }}
\end{equation}
and the phase frequency response can be written as
\begin{equation} \label{eq:bb}
\angle N\left( {B,\omega ,n} \right) \approx  - \arctan \left( {\frac{{m\omega }}{{\mu {B^{n - 1}}\Psi \left( n \right)}}} \right)
\end{equation}

Up to this point, we have obtained the frequency-domain response function of the system, which enables us to analyze the dynamic response performance of the system, such as bandwidth, settling time, and gain variation of SFC.

\subsection{Corollary 1: System Bandwidth}
Analyzing the system bandwidth of a robot controller helps evaluate its response speed and stability, determine the best parameters and design, and improve robot control performance and accuracy. The system bandwidth reflects the maximum frequency signal the controller can process, which is crucial for rapid response in robot applications. 

Consider a system of the form (\ref{eq:with_mass}), assuming $M>0, \mu>0, n>0$. Then the relationship between the input signal amplitude $A$ and the control bandwidth ${\omega_\text{c}}$ is given by 
\begin{equation} \label{eq:wc_A}
{\omega_\text{c}} \approx \frac{{{{\left( {\mu \Psi \left( n \right)} \right)}^{\frac{1}{n}}}}}{m}{\left( {\frac{{{A}}}{{\sqrt 2 }}} \right)^{\frac{{n - 1}}{n}}}
\end{equation}

\begin{proof}

Using equation (\ref{eq:AF}), we can express the steady-state amplitude-frequency response of the system at $\omega = 0$ and cutoff frequency $\omega = \omega_\text{c}$ as follows
\begin{equation} \label{eq:w0}
\left| {N\left( {B,0} \right)} \right| \approx \frac{1}{{\mu {B^{n - 1}}\Psi \left( n \right)}}
\end{equation}

\begin{equation} \label{eq:wb}
\left| {N\left( {B,{\omega_\text{c}}} \right)} \right| \approx \frac{1}{{\sqrt {{{\left( {m{\omega_\text{c}}} \right)}^2} + {{\left( {\mu {B^{n - 1}}\Psi \left( n \right)} \right)}^2}} }}
\end{equation}

The system bandwidth ${{\omega_\text{c}}}$ is typically defined as follows
\begin{equation} \label{eq:0.707}
    \left| {N\left( {B,{\omega_\text{c}}} \right)} \right| = \frac{{\sqrt 2 }}{2}\left| {N\left( {B,0} \right)} \right|
\end{equation}

Substituting equations (\ref{eq:w0}) and (\ref{eq:wb}) into equation (\ref{eq:0.707})
\begin{equation} \label{eq:bandwidth}
{\omega_\text{c}} \approx \frac{{\mu {B^n}\Psi \left( n \right)}}{{mB}}
\end{equation}

Substituting (\ref{eq:bandwidth}) into (\ref{eq:A(B,w)}), we get
\begin{equation} \label{eq:A_wc}
A(B,{\omega_\text{c}}) \approx \sqrt 2 \mu {B^n}\Psi \left( n \right)
\end{equation}

Using equations (\ref{eq:A_wc}) and (\ref{eq:bandwidth}), we can derive the relationship between the input signal amplitude $A$ and the control bandwidth as follows

\begin{equation} \label{eq:wc_A1}
{\omega_\text{c}} \approx \frac{{{{\left( {\mu \Psi \left( n \right)} \right)}^{\frac{1}{n}}}}}{m}{\left( {\frac{{{A}}}{{\sqrt 2 }}} \right)^{\frac{{n - 1}}{n}}}
\end{equation}

\end{proof}

\subsection{Corollary 2: System Time Constant}
The time constant represents the time required for the controller response to reach its peak value. In robot applications, the choice of time constant is crucial for ensuring that the controller can respond quickly and accurately to external disturbances, and achieve high-precision motion. 

Consider a system of the form (\ref{eq:with_mass}), assuming $M>0, \mu>0, n>0$. Then, the system time constant $\tau$ can be expressed as a function of $A$ as:
\begin{equation} \label{eq:TimeConstant}
\tau  \approx \frac{m}{{{{\left( {\mu \Psi \left( n \right)} \right)}^{\frac{1}{n}}}{A^{\frac{{n - 1}}{n}}}}}
\end{equation}

\begin{proof}

The time constant of a dynamic system is a characteristic parameter that defines the response time of the system to a step input. It represents the time it takes for the system's output to reach 63.2\% of its final value after a step input is applied. Taking a step signal as input, 
\begin{equation} \label{eq:step_A}
{f_\text{ext}}(t) = \left\{ {\begin{array}{*{20}{c}}
{0\begin{array}{*{20}{c}}
{}&{}&{}
\end{array}t = 0}\\
{{A_{{\rm{0}}}}\begin{array}{*{20}{c}}
{}&{}&{t > 0}
\end{array}}
\end{array}} \right.
\end{equation}

Then the frequency response of the output signal can be derived as

\begin{equation}
\begin{array}{*{20}{l}}
  {\dot x(j\omega ) = F(j\omega )N(B,j\omega )} \\ 
  { \approx \frac{{{A_{\text{0}}}}}{{j\omega }} \cdot \frac{B}{{\mu {B^n}\Psi \left( n \right) + jMB\omega }}} \\ 
  { = \left( {\frac{{{A_{\text{0}}}B}}{{\mu {B^n}\Psi \left( n \right)}}} \right)\left( {\frac{1}{{j\omega }} - \frac{{mB}}{{\mu {B^n}\Psi \left( n \right) + jMB\omega }}} \right)} 
\end{array}
\end{equation}

The time domain response can then be approximated as
\begin{equation}
\dot x(t) \approx \frac{{{A_{\text{0}}}B}}{{\mu {B^n}\Psi \left( n \right)}}\left( {1 - MB \cdot {e^{ - \frac{{\mu {B^n}\Psi \left( n \right)}}{{mB}}t}}} \right)
\end{equation}

Thus, the time constant of a dynamic system can be determined by

\begin{equation} \label{eq:tal}
\tau  \approx {\left. {\frac{{mB}}{{\mu {B^n}\Psi \left( n \right)}}} \right|_{\omega  = 0}}
\end{equation}

Based on equation~(\ref{eq:A(B,w)}), it can be inferred that $B$ is dependent on $A$. Once the system reaches a steady-state, it is possible to derive the following result
\begin{equation} \label{eq:B(A)}
B(A,0) \approx {\left( {\frac{A}{{\mu \Psi \left( n \right)}}} \right)^{\frac{1}{n}}}
\end{equation}

By substituting equation (\ref{eq:B(A)}) into equation (\ref{eq:tal}), the time constant of the system can be expressed as a function of $A$, specifically:
\begin{equation} \label{eq:TimeConstant_A}
\tau  \approx \frac{m}{{{{\left( {\mu \Psi \left( n \right)} \right)}^{\frac{1}{n}}}{A^{\frac{{n - 1}}{n}}}}}
\end{equation}

And the system settling time is
\begin{equation} \label{eq:Tss}
{T_{{\text{ss}}}} \approx 4\tau  = \frac{{4M}}{{{\mu ^{\frac{1}{n}}}{A^{\frac{{n - 1}}{n}}}}}
\end{equation}  

\end{proof}

From the above equation we can know what the system settling time depends on, so it can be customized by choosing appropriate values for the $\mu,M,n$ parameters in the above equation.

\subsection{ Corollary 3: Correlation between Variation of System
Gain and Power-law} 

Analyzing the correlation between the variation of system gain and power-law in SFC systems can help in selecting the optimal power-law parameter to meet the system's dynamic response requirements and prevent unreasonable power-law parameters that can make the controller close to ill-condition.

We examine the system gain at steady-state after the controller is subjected to a step signal excitation. We use a step signal (\ref{eq:step_A}) as the input and substitute (\ref{eq:B(A)}) into (\ref{eq:w0}) to obtain the expression (\ref{eq:N_A0}).
\begin{equation} \label{eq:N_A0}
\left| {N\left( {A,0,n} \right)} \right| \approx {\left( {\mu \Psi \left( n \right)} \right)^{ - \frac{1}{n}}}{A^{\frac{{1 - n}}{n}}}
\end{equation}

The system gain is usually represented in the unit of decibels (dB) in Bode diagrams for ease of analysis, and can be expressed as
\begin{equation}
Q\left( {A,0,n} \right) = 20\log \left( {\left| {N\left( {A,0,n} \right)} \right|} \right)
\end{equation}
 
The system gain change $\Delta {Q_m}$ is defined as the difference in system gain between the input signal amplitudes of ${A_0}$ and ${A_m} = {10^m}{A_0}$ (where $m$ is a positive integer), i.e.,
\begin{equation} \label{eq:dQ}
    \Delta {Q_m} = Q({A_m},0,n) - Q({A_0},0,n)
\end{equation}

Consider a nonlinear SFC system of the form (\ref{eq:with_mass}) with $n>1$, $M>0$ and $\mu>0$. The system gain change of this nonlinear system ($n>1$) is negative, indicating that as the input signal amplitude increases, the attenuation effect becomes more pronounced. Moreover, the variation of system gain $\Delta {Q_m}$ is solely dependent on the power $n$, and there exists a lower limit. Specifically,
\begin{equation} \label{eq:lim}
\mathop {\lim }\limits_{n \to \infty } \Delta {Q_m} = - 20m
\end{equation}

\begin{proof}
    We start with the expression for the change of system gain given in equation (\ref{eq:dQ}).
\begin{equation*}
\Delta {Q_m} = Q\left( {{A_m},0,n} \right) - Q\left( {{A_0},0,n} \right) = 20m\frac{{1 - n}}{n}
\end{equation*}

Next, we take the limit of $\Delta {Q_m}$ as $n$ approaches infinity.
\begin{align*}
\mathop {\lim }\limits_{n \to \infty } \Delta {Q_m} = \mathop {\lim }\limits_{n \to \infty } \left( {20m\frac{{1 - n}}{n}} \right)\
= -20m
\end{align*}

\end{proof}
Thus, we have shown that when $n$ tends to infinity, the limit of the system gain change $\Delta {Q_m}$ is equal to $-20m$, as given in equation (\ref{eq:lim}). This shows that the system attenuation does not decay indefinitely as the SFC increases by $n$, and that the choice of $n$ does not need to be large but only sufficient to meet the requirements.

\section{Appendix E: Discrete control constraints}
The aforementioned analysis was conducted based on the assumption of a continuous system, whereas in reality, robot systems are often digitally discrete. At high control frequencies, the discrete system can be approximated as a continuous system. Nevertheless, when the control frequency is low, the control constraints associated with discretization must be taken into account. The Laplace transform is a valuable approach for dealing with discrete linear systems, but its application to nonlinear discrete systems can be problematic. This section focuses on analyzing the convergence acceleration of the nonlinear SFC directly from the discretized recursion. Such an analysis facilitates the derivation of parameter constraints that ensure stability of the discrete system.

 Consider a discretized SFC system of the form (\ref{eq:Discrete}) with parameters $n>0$, $M>0$, and $\mu>0$. Let ${f_\text{ext,max}}$ be the maximum value of external input force. Then, the upper limit of the sample time is constrained by
\begin{equation} \label{eq:constraint2}
    \Delta T < 2M{\mu ^{\frac{{ - 1}}{n}}}{n^{ - 1}}{\left| {{f_\text{ext,max}}} \right|^{\frac{{1 - n}}{n}}}
\end{equation}

Furthermore, the relationship between the system bandwidth ${\omega_\text{c}}$ and the sample time under the discretized SFC is constrained by
\begin{equation} \label{eq:wwc2}
{\omega_\text{c}} < {2^{\frac{{n + 1}}{{2n}}}}\frac{{\Psi {{\left( n \right)}^{\frac{1}{n}}}}}{{\Delta Tn}}{\left| {\frac{{{f_\text{ext}}}}{{{f_{\text{ext},\max }}}}} \right|^{\frac{{n - 1}}{n}}}
\end{equation}

\begin{proof}

Calculate the acceleration value at time $t+1$.
\begin{equation} \label{eq:at+}
\ddot x\left( {t + 1} \right) = {m^{ - 1}}\left( {{f_\text{ext}}\left( {t + 1} \right) - \mu {{\left| {\dot x\left( t \right)} \right|}^{n - 1}}\dot x\left( t \right)} \right)
\end{equation}

Let $\Delta U\left( t \right)$ denote the change in the damping term.
\begin{equation} \label{eq:U}
\Delta U\left( t \right) = \mu {\left| {\dot x\left( t \right)} \right|^{n - 1}}\dot x\left( t \right) - \mu {\left| {\dot x\left( {t - 1} \right)} \right|^{n - 1}}\dot x\left( {t - 1} \right)
\end{equation}

Subtract (\ref{eq:Discrete}) from (\ref{eq:at+}) and substitute (\ref{eq:U}) into the equation.
\begin{equation}\label{eq:a-a}
\begin{gathered}
  \ddot x\left( t \right) - \ddot x\left( {t + 1} \right) \hfill \\
   = {m^{ - 1}}\left( {{f_\text{ext}}\left( t \right) - {f_\text{ext}}\left( {t + 1} \right) + \Delta U\left( t \right)} \right) \hfill \\ 
\end{gathered} 
\end{equation}

Calculate $\mu {\left| {\dot x\left( t \right)} \right|^{n - 1}}\dot x\left( t \right)$ iteratively through (\ref{eq:Discrete})
\begin{equation} \label{eq:ft+}
\begin{gathered}
  \mu {\left| {\dot x\left( t \right)} \right|^{n - 1}}\dot x\left( t \right) \hfill \\
   = \mu {\left| {\dot x\left( {t - 1} \right) + \ddot x\left( t \right)\Delta T} \right|^n}{\text{sign}}\left( {\dot x\left( {t - 1} \right) + \ddot x\left( t \right)\Delta T} \right) \hfill \\ 
\end{gathered} 
\end{equation}

Substitute (\ref{eq:ft+}) into (\ref{eq:U}) and calculate the approximate value by Binomial Theorem:
\begin{equation} \label{eq:f-f}
\Delta U\left( t \right) \approx \mu n\Delta T{\left| {\dot x\left( {t - 1} \right)} \right|^{n - 1}}\ddot x\left( t \right)
\end{equation}

The external force is assumed to be zero or constant in the analysis. Thus, the term ${{f_\text{ext}}\left( t \right) - {f_\text{ext}}\left( {t + 1} \right)}$ is zero. Substitute (\ref{eq:f-f}) into (\ref{eq:a-a}) 
\begin{equation}
    \ddot x\left( t \right) - \ddot x\left( {t + 1} \right) \approx {m^{ - 1}}\mu n\Delta T{\left| {\dot x\left( {t - 1} \right)} \right|^{n - 1}}\ddot x\left( t \right)
\end{equation}

Then the relationship between $\ddot x\left( {t + 1} \right)$ and $\ddot x\left( {t} \right)$ can be obtained
\begin{equation}
\left( {1 - \mu {m^{ - 1}}\Delta Tn{{\left| {\dot x\left( {t - 1} \right)} \right|}^{n - 1}}} \right)\ddot x\left( t \right) \approx \ddot x\left( {t + 1} \right)
\end{equation}

In order to ensure acceleration convergence, the following constraint need to be satisfied
\begin{equation} \label{eq:constraint3}
\left| {1 - \mu {m^{ - 1}}\Delta Tn{{\left| {\dot x\left( {t - 1} \right)} \right|}^{n - 1}}} \right| < 1
\end{equation}

Therefore, there is an upper limit of the sample time $\Delta T$.
\begin{equation} \label{eq:T_xd}
    \Delta T < 2M{\left( {\mu n} \right)^{ - 1}}\left| {\dot x} \right|_{\max }^{1 - n}
\end{equation}

When the controller (\ref{eq:Discrete}) reaches a steady state, the acceleration is zero, and the velocity output reaches its maximum value.
\begin{equation} \label{eq:xd_max}
    {\left| {\dot x} \right|_{\max }} = {\left( {\frac{{\left| {{f_\text{ext,max}}} \right|}}{\mu }} \right)^{\frac{1}{n}}}
\end{equation}

According to (\ref{eq:T_xd}) and  (\ref{eq:xd_max}), the upper limits of the sample time can be rewritten as 

\begin{equation} \label{eq:constraint4}
    \Delta T < 2M{\mu ^{\frac{{ - 1}}{n}}}{n^{ - 1}}{\left| {{f_\text{ext,max}}} \right|^{\frac{{1 - n}}{n}}}
\end{equation}

Since $n>1$, the upper limit of the sample time is negatively related to the external force amplitude. Substituting equation (\ref{eq:wc_A}) into (\ref{eq:constraint4}), the constrained relationship between the system bandwidth and sampling time under SFC can be calculated as 

\begin{equation} \label{eq:wwc3}
{\omega_\text{c}} < {2^{\frac{{n + 1}}{{2n}}}}\frac{{\Psi {{\left( n \right)}^{\frac{1}{n}}}}}{{\Delta Tn}}{\left| {\frac{{{f_\text{ext}}}}{{{f_{\text{ext},\max }}}}} \right|^{\frac{{n - 1}}{n}}}
\end{equation}

\end{proof}

\section{Appendix F: Parameter auto-tuning algorithm}
Here, we outline the purpose and design rationale behind each step of the controller parameter auto-tuning Algorithm~\ref{alg:auto}.

\textbf{Input: } Determine the dynamic responsiveness requirements in human-robot interaction, as shown in Table~\ref{tab:require}, and input them as known parameters.
\begin{center}
${f_{{\text{ease}}}},{f_{{\text{interf}}}},{{\dot {\bar x}}_\text{d}},{{\dot {\bar x}}_c},{\omega _\text{c\_ease}}, \Delta T$
\end{center}

\textbf{Step 1: (Line 1) }Compute the power exponent $n$ that satisfies the
given requirement. The equation (\ref{eq:A(B,w)}) depicts the correlation between force and velocity across varying frequencies and amplitudes. The maximum gain occurs at lower frequencies, where ${\left( {mB\omega } \right)^2} \ll {\left( {\mu {B^n}\Psi \left( {{n^{ - 1}}} \right)} \right)^2}$. As a result, the equation (\ref{eq:A(B,w)}) can be simplified to
\begin{equation} \label{eq:MNNFC_A_LowFrequency}
A = \Psi \left( {{n^{ - 1}}} \right)\mu B_{\max }^n
\end{equation}
where, A represents the magnitude of force and B denotes the maximum amplitude of velocity at that point.

In the coupled state, we desire the robot to move compliantly with the force $f_\text{ease}$ at a velocity of $\dot {\bar x}_d$. Equation (\ref{eq:MNNFC_A_LowFrequency}) indicates that:
\begin{equation} \label{eq:ease}
    {f_\text{ease}} = \Psi \left( {{n^{ - 1}}} \right)\mu {\left( {\frac{{{{\dot {\bar x}}_\text{d}}}}{g}} \right)^n}
\end{equation}

At the same time, if an impact force occurs, we aim to limit the maximum speed jump to no more than $\dot {\bar x}_c$. Therefore, we have the following inequality:
\begin{equation} \label{eq:interf}
f_{{\text{interf}}} \leqslant \Psi \left( {{n^{ - 1}}} \right)\mu {\left( {\frac{{{{\dot {\bar x}}_c}}}{g}} \right)^n}
\end{equation}

To satisfy equations (\ref{eq:ease}) and (\ref{eq:interf}), $n$ must adhere to the following constraint:
\begin{equation} \label{eq:n}
    n \geq {{\log \left| {\frac{{{f_{{\text{interf}}}}}}{{{f_{{\text{ease}}}}}}} \right|} \mathord{\left/
 {\vphantom {{\log \left| {\frac{{{f_{{\text{interf}}}}}}{{{f_{{\text{ease}}}}}}} \right|} {\log \left| {\frac{{{{\dot {\bar x}}_c}}}{{{{\dot {\bar x}}_\text{d}}}}} \right|}}} \right.
 \kern-\nulldelimiterspace} {\log \left| {\frac{{{{\dot {\bar x}}_c}}}{{{{\dot {\bar x}}_\text{d}}}}} \right|}}
\end{equation}

According to Equation (\ref{eq:dQ}), it is sufficient to choose a value of $n$ that meets the requirement, rather than choosing a larger value. Therefore, we select the minimum value of $n$ that satisfies inequality (\ref{eq:n}). 

\textbf{Step 2: (Line 2) } Determining the system bandwidth $\omega_{\text{c}\_\max}$ of the SFC under impact interference, while ensuring it meets the requirement for smooth traction functionality. The system bandwidths under impact force, $f_{\text{interf}}$, and traction force, $f_{\text{ease}}$, can be estimated using Equation~\ref{eq:wc_A}:
\begin{equation} \label{eq:w_ease}
{\omega _{\text{c}\_\max}} \approx \frac{{{{\left( {\mu \Psi \left( n \right)} \right)}^{\frac{1}{n}}}}}{m}{\left( {\frac{{{f_{{\text{interf}}}}}}{{\sqrt 2 }}} \right)^{\frac{{n - 1}}{n}}}
\end{equation}
\begin{equation} \label{eq:w_interf}
{\omega _\text{c\_ease}} \approx \frac{{{{\left( {\mu \Psi \left( n \right)} \right)}^{\frac{1}{n}}}}}{m}{\left( {\frac{{{f_\text{ease}}}}{{\sqrt 2 }}} \right)^{\frac{{n - 1}}{n}}}
\end{equation}

Although the specific parameters of the controller are unknown, the maximum traction force $f_{\text{ease}}$, the maximum impact force $f_{\text{interf}}$, and the desired system bandwidth $\omega_{\text{c\_ease}}$ as specified in the control requirements can be used with Equation (\ref{eq:w_ease}) and (\ref{eq:w_interf}) to determine the maximum system bandwidth under impact conditions
\begin{equation} \label{eq:w_cmax}
    {\omega _{\text{c}\_\max}} = {\omega _\text{c\_ease}}{\left( {\frac{{{f_{{\text{interf}}}}}}{{{f_{{\text{ease}}}}}}} \right)^{\frac{{n - 1}}{n}}}
\end{equation}

\textbf{Step 3: (Line 3-5)}  Evaluate the system bandwidth to determine if it meets the constraint requirements, and make necessary modifications to address any deviations from the constraints. Based on the discrete constraints (\ref{eq:constraint4}), it is determined that the system parameters need to meet the requirements under impact disturbances.
\begin{equation} \label{eq:dT_Fmax}
    \Delta T < 2M{\mu ^{\frac{{ - 1}}{n}}}{n^{ - 1}}{\left| {{f_{{\text{interf}}}}} \right|^{\frac{{1 - n}}{n}}}
\end{equation}

By substituting equation (\ref{eq:w_interf}) into equation (\ref{eq:dT_Fmax}), the constraint on the system bandwidth under impact disturbance can be formulated as:
\begin{equation} \label{eq:w_constraint}
    {\omega _{\text{c}\_\max}} \leqslant \frac{1}{{\Delta Tn}}{2^{\frac{{1 + n}}{{2n}}}}
\end{equation}

If the value of ${\omega _{\text{c}\_\max}}$ obtained from equation (\ref{eq:w_cmax}) does not satisfy the constraint (\ref{eq:w_constraint}), it may be necessary to adjust the desired system bandwidth during the traction state. Let ${\omega _{\text{c}\_\max}}$ be the upper limit that satisfies constraint (\ref{eq:w_constraint}). The adjusted desired bandwidth can be obtained using Equation~(\ref{eq:w_cmax}).
\begin{equation}
    {\omega _\text{c\_ease}} = \frac{2}{{\Delta Tn}}{\left( {\frac{{{f_{{\text{ease}}}}}}{{\sqrt 2 {f_{{\text{interf}}}}}}} \right)^{\frac{{n - 1}}{n}}}
\end{equation}

\textbf{Step 4: (Line 6)} Select the virtual inertia of the system and calculate the apparent damping that meets the bandwidth requirements. According to (\ref{eq:wc_A}), the relative relationship between $\mu$ and $M$ determines the bandwidth of the system. To facilitate analysis, a normalized inertia
parameter $M = 1$ is chosen. Then calculate the parameter $\mu$ according to the equation (\ref{eq:w_interf}) as follows
\begin{equation}
\mu  \approx \frac{{{{\left( {m{\omega _\text{c\_ease}}} \right)}^n}}}{{\Psi \left( n \right)}}{\left( {\frac{{\sqrt 2 }}{{{f_\text{ease}}}}} \right)^{n - 1}}
\end{equation}

\textbf{Step 5: (Line 7)} Calculating the coefficient for adjusting the gain of the system.

 By adding a gain term $g$ after (\ref{eq:with_mass}), the amplitude frequency response curve can be translated longitudinally as a whole in the Bode plot. In order to make the output during compliant drag no greater than ${{\dot {\bar x}}_\text{d}}$, set the gain $g$ as follows
\begin{equation}
    g = {{\dot {\bar x}}_\text{d}}{\left( {\frac{\mu }{{{f_{{\text{ease}}}}}}} \right)^{\frac{1}{n}}}
\end{equation}

\textbf{Output: }Obtain controller parameters that ultimately fulfill the requirements for interactivity.
\begin{center}
$\mu, n, g$
\end{center}

To sum up, we get the parameter setting steps of SFC, as shown in the Algorithm~\ref{alg:auto}.

\section{Appendix G: Ill-condition caused by excessively high power law}

Ill-conditioning refers to the system's response being highly sensitive to small changes in inputs or parameters, resulting in instability in control. In such cases, even small variations in input signals or parameters can cause significant fluctuations or unpredictable behavior in the system's output. We analyze the discretized implementation of SFC in practical use. In equation (20), the velocity at the previous time step, ${\dot x(t - 1)}$, is used as a parameter to calculate the acceleration, ${\ddot x(t)}$, for the next VDC cycle. Assuming the true value of velocity at the previous cycle is $\dot x{(t - 1)_\text{real}} = 1$, and there exists a small perturbation of $\delta  = 0.1$, which leads to a perturbed velocity of $\dot x{(t - 1)_\text{real}} + \delta  = 1.1$. For the sake of simplicity in analysis, let's assume other parameters are set as $M = \mu =1 $, and ${f_{{\rm{ext }}}}(t) = 10$. Based on equation (20), we can obtain the corresponding true VDC acceleration and perturbed VDC acceleration as $\ddot x{(n)_\text{real}} = 10 - {1^n}$, $\ddot x{(n)_\text{dist}} = 10 - {1.1^n}$.
For a lower power-law parameter of the controller ($n=3$), the calculated values are $\ddot x{(3)_\text{real}} = 10 - {1^3} = 9,\ddot x{(3)_\text{dist}} = 10 - {1.1^3} = 8.67$. In this case, the impact of the small perturbation in the parameter on VDC acceleration is not significant. However, for a higher power-law parameter of the controller ($n=100$), the calculated values are $\ddot x{(100)_\text{real}} = 10 - {1^{100}} = 9$, $\ddot x{(100)_\text{dist}} = 10 - {1.1^{100}} =  - 13770.61$. Compared to the case of $n=3$f, the VDC acceleration undergoes a tremendous disturbance.
Hence, it can be demonstrated that when the power-law parameter of the controller is too high, it can lead to the system being in an ill-conditioned state. The choice of n should not necessarily be a very large value, but rather just sufficient to meet the necessary requirements.

\section{Appendix H: SFC Implementation Details}
\begin{figure}[!t]
\centering
\includegraphics[width=3.3in]{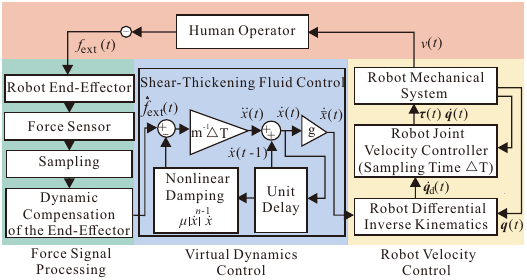}%
\caption{The interconnection and composition of each component in the human-robot coupled state}
\label{fig:real_frame}
\end{figure}

The implementation of SFC in pHRI consists of four main components: the human operator, the force signal processing, the virtual dynamics control, and the robot velocity control. The interconnection and composition of each component in the human-robot coupled state are shown in Figure \ref{fig:real_frame}. 

1)	The Human Operator: Drawing inspiration from the works of \cite{colgate1997passivity,keemink2018admittance}, we model the human operator as a pure impedance system composed of a mass $m_\text{h}>0$ and damping $b_\text{h}>0$.The transfer function of the human operator is given by
\begin{equation} \label{eq:humanZh}
    {Z_{\rm{h}}}\left( s \right) = \frac{1}{{{m_\text{h}}s + {b_\text{h}}}}
\end{equation}

\begin{figure}[!t]
\centering
\includegraphics[width=2.8in]{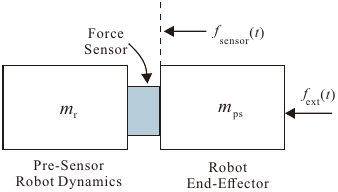}%
\caption{Schematic diagram of interactive force measurement.}
\label{fig:Force_sensor}
\end{figure}

2)	 Force Signal Processing: This part is for measuring the contact forces between the human and the robot, as well as compensating for the dynamic disturbances of the end-effector. Specifically, the contact force exerted by the human operator is applied to the robot's end-effector (e.g., robot gripper). It is then sensed by a six-axis force/torque sensor located at the wrist, which converts it into a digital signal through sampling for computer processing. The robot's end-effector is typically modeled as a mass block ($m_\text{ps}$) fixedly connected to the force sensor (\cite{keemink2018admittance}). As shown in the Figure \ref{fig:Force_sensor}.

\begin{figure}[!t]
\centering
\includegraphics[width=2 in]{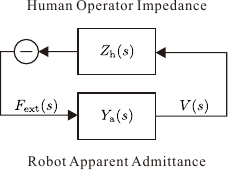}%
\caption{The interconnection between an admittance-controlled robot, characterized by its apparent admittance $Y_\text{a}$, and a human with an impedance $Z_h$, results in a closed-loop (coupled) interaction behavior.}
\label{fig:ZY_transform}
\end{figure}

Due to the presence of the end-effector mass ($m_\text{ps}$) , the force sensed by the force sensor ${f_\text{sensor}}(t)$ comprises not only the external force ${f_\text{ext}}(t)$ but also the dynamic disturbance force exerted by $m_\text{ps}$, denoted as $\delta \left( {{m_\text{ps}},t} \right)$. 
\begin{equation}
    {f_\text{sensor}}\left( t \right) = {f_\text{ext}}\left( t \right) + \delta \left( {{m_\text{ps}},t} \right)
\end{equation}
In order to recover the real external forces, we computed the dynamic compensation of the end-effector. The studies conducted by \cite{landi2016tool,yu2021bias} provide useful references for the methodology. The estimated value of the external force is obtained as
\begin{equation}
    {\hat f_\text{ext}}(t) = {f_\text{sensor}}(t) - \delta({m_\text{ps}},t)
\end{equation}
In subsequent analysis, we assume the dynamic compensation of the end-effector to be accurate, implying that
\begin{equation}
    {\hat f_\text{ext}}(t) = f_\text{ext}(t)
\end{equation}

Additionally, in the real world experiments, we utilized the ATI-Delta330 force sensor with a sampling frequency of 1.5 kHz. The frequency of contact forces between humans and robots is lower than 100 Hz. Due to the significantly higher sampling frequency compared to the frequency of the actual signals, accurate reconstruction is possible. Therefore, subsequent modeling analysis ignores the influence of sampling.

3)	Virtual Dynamics Control: This module utilizes SFC to impart the dynamic characteristics of shear-thickening fluids to the robot. Analyses in the previous paper have mostly focused on continuous ideal models. Here, modeling and analysis are performed with a practical discrete model. Similar to the approaches utilized by \cite{ferraguti2019variable,kang2019variable}, the forward Euler method is used in this paper for discrete integration calculations (\ref{eq:Discrete}).  The primary objective of this module is to calculate the mapping from contact forces to robot operational space velocity commands. 

Please note that the sampling time $\Delta T$ of the system here is consistent with the subsequent velocity controller period, as the forward Euler integration in (\ref{eq:Discrete}) essentially estimates the state update of the robot within the $\Delta T$ velocity control cycle. 
Figure \ref{fig:real_frame} showcases the controller block diagram, incorporating the utilization of a Unit Delay (${H_{{\rm{ud}}}}(s){\rm{ = }}{e^{ - \Delta Ts}}$) to maintain the velocity and subsequently apply it in the subsequent time step.

4) Robot Velocity Control: This module primarily focuses on controlling the robot to achieve the desired operational space velocity. When a velocity command in Cartesian space is received, it is mapped to joint space using the Jacobian matrix, resulting in the desired joint space velocity ${\dot q_d}$. 
The robot we utilize is a non-backdrivable robot, which has a joint space velocity control interface for external control. The joint space velocity controller is commonly constructed using feedforward compensation based on robot dynamics and PD feedback that relies on joint velocity error. This controller allows for precise joint velocity control with a control period of $\Delta T$ .
 In the subsequent analysis, we assume that the velocity control of the inner loop is sufficiently accurate. Additionally, due to its control period of $\Delta T$, the Robot Velocity Control exhibits an overall Zero-Order Hold (ZOH) dynamic characteristic when observed externally.
 
\begin{equation}
    {H_{{\rm{zoh}}}}(s){\rm{ = }}\frac{{1 - {e^{ - \Delta Ts}}}}{{\Delta T{\rm{s}}}}
\end{equation}

\section{Appendix I:Coupled Stability Analysis}

\begin{figure}[!t]
\centering
\includegraphics[width=3in]{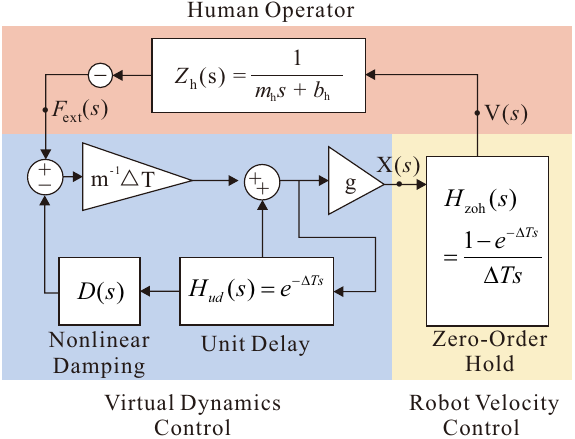}%
\caption{A simpified model of the entire system.}
\label{fig:real_transform}
\end{figure}

Based on the Appendix H analysis, we can simplify the model of the entire system as shown in Figure \ref{fig:real_transform}. The system essentially consists of a human with impedance characteristics (${Z_\text{a}}$) and a robot with admittance characteristics (${Y_\text{a}}$), as illustrated in Figure \ref{fig:ZY_transform}. The Virtual Dynamics Control and Robot Velocity Control modules depicted in Figure \ref{fig:real_transform} collectively form the apparent admittance ( ${Y_\text{a}}$) of the robot in Figure \ref{fig:ZY_transform}. This configuration conforms to the basic model of pHRI. 

\begin{proof}
Figures \ref{fig:real_transform} and \ref{fig:ZY_transform} illustrate the connection between the two systems via a power-continuous coupling. The phase property of passive systems restricts the overall phase of the \textit{open-loop} transfer function to be within the range of -180° to +180° (the phase of each individual port function is between -90° and +90°, and they are summed together). As the phase never exceeds these limits, the coupled system remains stable and at worst marginally stable (\cite{colgate1988robust}). 

This finding suggests that in human-robot interaction, if humans are considered passive, as long as the apparent admittance phase of the robot ($Y_\text{a}$) does not surpass ±90°, the coupled system remains stable. Equation (\ref{eq:humanZh}) models the human as a mass-damper system (${m_\text{h}} > 0,{b_\text{h}} > 0$), indicating that humans are passive. Next, we analyze the phase shift characteristics of the robot's apparent admittance $Y_\text{a}$. In Figures \ref{fig:real_transform}, the transfer function of the robot's apparent admittance can be expressed as follows

\begin{equation} \label{eq:Yas}
{Y_\text{a}}(s) = \frac{{V(s)}}{{F(s)}} = \frac{{V(s)}}{{X(s)}}\frac{{X(s)}}{{{F_{{\rm{ext}}}}(s)}}
\end{equation}

\begin{equation} \label{eq:Vs}
\frac{{V(s)}}{{X(s)}} = \frac{{{m^{ - 1}}\Delta T}}{{1 + {H_{{\rm{ud}}}}(s)\left( {D(s){m^{ - 1}}\Delta T - 1} \right)}}
\end{equation}

\begin{equation} \label{eq:Xs}
\frac{{X(s)}}{{{F_{{\rm{ext}}}}(s)}} = {H_{{\rm{zoh}}}}(s)
\end{equation}

In this expression, $D(s)$ corresponds to the nonlinear damping term, which cannot be directly solved through Laplace transformation. However, it can be analyzed through the approximation of describing functions (\cite{slotine1991applied}). The specific analysis procedure aligns with Appendix D. According to (\ref{eq:b1}), when $\dot x\left( t \right) = B\sin (\omega t)$, the response of the nonlinear damping term is as follows
\begin{equation}
    D\left( {\dot x\left( t \right)} \right) \approx \mu {B^n}\Psi \left( n \right)\sin (\omega t)
\end{equation}
Thus, the describing function of $D(t)$ can be represented as
\begin{equation}
    D(j\omega ) \approx \frac{{\mu {B^n}\Psi \left( n \right)\sin (\omega t)}}{{B\sin (\omega t)}} = \mu {B^{n - 1}}\Psi \left( n \right)
\end{equation}
Substituting $s = j\omega$, ${H_{{\rm{zoh}}}}(s)$, ${H_{{\rm{ud}}}}(s)$, and $D(j\omega )$ into (\ref{eq:Vs}) and (\ref{eq:Xs}) yields
\begin{equation} \label{eq:Vs_j}
    \frac{{V(j\omega )}}{{X(j\omega )}} = \frac{{{m^{ - 1}}\Delta T}}{{1 + \left( {Q - 1} \right)\cos \left( {\Delta T\omega } \right) - j\left( {Q - 1} \right)\sin \left( {\Delta T\omega } \right)}}
\end{equation}
\begin{equation} \label{eq:Xs_j}
    \frac{{X(j\omega )}}{{{F_\text{ext}}(j\omega )}} = \frac{{\sin \left( {\Delta T\omega } \right) - j\left( {1 - \cos \left( {\Delta T\omega } \right)} \right)}}{{\omega \Delta T}}
\end{equation}
Based on (\ref{eq:Vs_j}) and (\ref{eq:Xs_j}), we can determine the phase shift of each module to be
\begin{equation} \label{eq:Vs_phase}
\angle \frac{{V(j\omega )}}{{X(j\omega )}} = \arctan \left( {\frac{{\left( {Q - 1} \right)\sin \left( {\Delta T\omega } \right)}}{{1 + \left( {Q - 1} \right)\cos \left( {\Delta T\omega } \right)}}} \right)
\end{equation}

\begin{equation} \label{eq:Xs_phase}
\angle \frac{{X(j\omega )}}{{{F_\text{ext}}(j\omega )}} = \arctan \left( {\frac{{\cos \left( {\Delta T\omega } \right) - 1}}{{\sin \left( {\Delta T\omega } \right)}}} \right) =  - \frac{{\Delta T\omega }}{2}
\end{equation}

If $0 < Q < 1$ and $0 < \Delta T\omega  < \pi$, it follows that 

\begin{equation}  \label{eq:Vs_phase2}
\left\{ \begin{array}{l}
\angle \frac{{V(j\omega )}}{{X(j\omega )}} > \arctan \left( {\frac{{ - \sin \left( {\Delta T\omega } \right)}}{{1 - \cos \left( {\Delta T\omega } \right)}}} \right) = \frac{{\Delta T\omega }}{2} - \frac{\pi }{2}\\
\angle \frac{{V(j\omega )}}{{X(j\omega )}} < \arctan \left( {\frac{{0 \cdot \sin \left( {\Delta T\omega } \right)}}{{1 + 0 \cdot \cos \left( {\Delta T\omega } \right)}}} \right) = 0
\end{array} \right.
\end{equation}

According to (\ref{eq:Xs_phase}) and (\ref{eq:Vs_phase2}), the following inferences can be made
\begin{equation}
    - \frac{{\Delta T\omega }}{2} > \left( {\angle \frac{{V(j\omega )}}{{X(j\omega )}} + \angle \frac{{X(j\omega )}}{{{F_\text{ext}}(j\omega )}}} \right) >  - \frac{\pi }{2}
\end{equation}

Since $\angle {Y_\text{a}} = \angle \frac{{V(j\omega )}}{{X(j\omega )}} + \angle \frac{{X(j\omega )}}{{{F_\text{ext}}(j\omega )}}$, it can be inferred that

\begin{equation}
    \angle {Y_\text{a}} \in \left( { - \frac{\pi }{2}, - \frac{{\Delta T\omega }}{2}} \right)
\end{equation}

As $0 < \Delta T\omega  < \pi$, we can conclude that
\begin{equation}
    \left| {\angle {Y_\text{a}}} \right| < \frac{\pi }{2}
\end{equation}
The phase shift of the apparent admittance ${Y_\text{a}}$ of the robot is less than ±90 degrees, indicating that the pHRI system under SFC control, as shown in Figure \ref{fig:real_transform}, satisfies the requirement of coupled stability.

\end{proof}

\end{document}